# HyperFree: A Channel-adaptive and Tuning-free Foundation Model for Hyperspectral Remote Sensing Imagery

Jingtao Li[*1], Yingyi Liu[*1], Xinyu Wang[1†], Yunning Peng[1], Chen Sun[1], Shaoyu Wang[2], Zhendong Sun[1], Tian Ke[1], Xiao Jiang[1], Tangwei Lu[1], Anran Zhao[1], Yanfei Zhong[1†]
[1]Wuhan University [2]Seoul National University
{Jintaoli,liuyingyi,wangxinyu,pengyunning,sunchen,
sunzhendong,ketian,jiangxiao,lutangwei,anranzhao415,zhongyanfei}@whu.edu.cn
wsy1995@snu.ac.kr

## Abstract

*Advanced interpretation of hyperspectral remote sensing images benefits many precise Earth observation tasks. Recently, visual foundation models have promoted the remote sensing interpretation but concentrating on RGB and multispectral images. Due to the varied hyperspectral channels, existing foundation models would face image-by-image tuning situation, imposing great pressure on hardware and time resources. In this paper, we propose a tuning-free hyperspectral foundation model called HyperFree, by adapting the existing visual prompt engineering. To process varied channel numbers, we design a learned weight dictionary covering full-spectrum from* $0.4 \sim 2.5\,\mu m$*, supporting to build the embedding layer dynamically. To make the prompt design more tractable, HyperFree can generate multiple semantic-aware masks for one prompt by treating feature distance as semantic-similarity. After pre-training HyperFree on constructed large-scale high-resolution hyperspectral images, HyperFree (1 prompt) has shown comparable results with specialized models (5 shots) on 5 tasks and 11 datasets. Code and dataset are accessible at https://rsidea.whu.edu.cn/hyperfree.htm.*

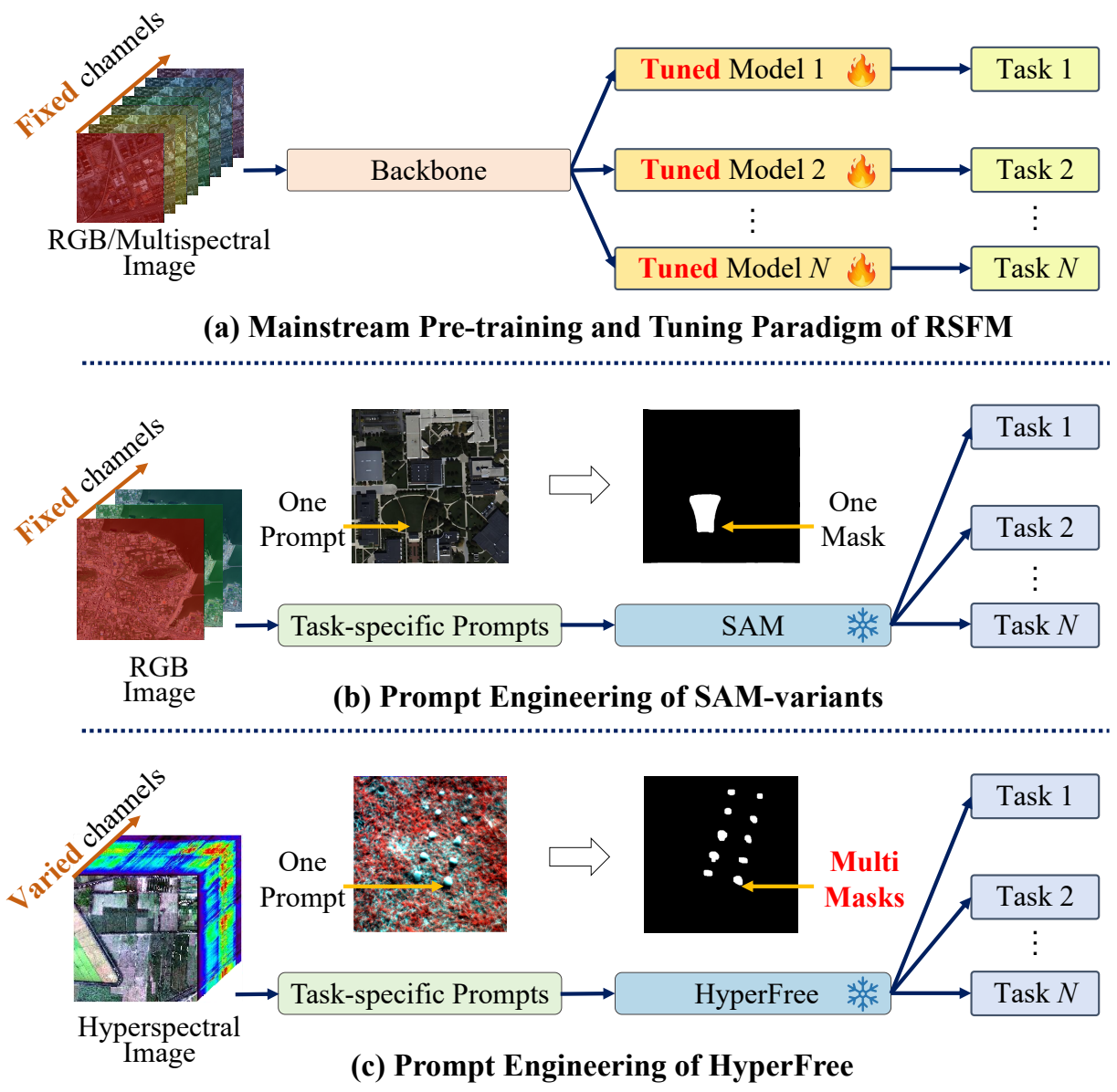

Figure 1. Comparison of existing RSFMs with the proposed HyperFree model. (a) The pre-training and tuning paradigm is the mainstream approach, where fine-tuning is essential for handling unseen HSIs. (b) The prompt engineering paradigm reduces the tuning burden but struggles to handle HSIs with hundreds of spectral channels. (c) HyperFree can directly process unseen HSIs with varying channel counts without fine-tuning, and it can identify multiple masks of the same class using only a single prompt.

## 1. Introduction

Hyperspectral imaging has been recognized by the European Union as one of the Top 100 Disruptive Technologies [58], due to its ability to provide high-resolution spectral data for precise Earth observation across diverse applications [64], including environmental monitoring [31], agriculture [69], and defense [20]. Unlike traditional RGB or multispectral imagery, Hyperspectral imagery (HSI) offers several unique features: high spectral resolution (typically 10 nm), an extensive wavelength range (400–2500 nm), and dozens to hundreds of spectral channels [64], which enable the detailed identification of diverse targets, while also increasing the complexity of data processing and analysis.

Recently, remote sensing foundation models (RSFMs) have demonstrated advanced interpretive capabilities, primarily focusing on multispectral and RGB images [4, 7, 13, 25, 38, 49], while also opening new possibilities for analyzing hyperspectral images. RSFMs follow two main paradigms. The predominant paradigm is pre-training and tuning (P-T), illustrated in Figure 1(a), which begins with training a large-scale backbone model in a self-supervised manner, such as masked autoencoders (MAE) [12], followed by fine-tuning with task-specific heads for various downstream applications [7, 13, 38, 49]. In the P-T paradigm, both the pre-training and the fine-tuning stages can be time-consuming. Second paradigm is prompt engineering (P-E), as illustrated in Figure 1(b), which focuses on designing task-specific prompts for a frozen Segment Anything Model (SAM) [17] to directly perform downstream tasks without additional model tuning [4, 25], making it more efficient and user-friendly in practical.

*Equal contribution.
†Corresponding authors.

The P-E paradigm is particularly well-suited for processing HSI, as it can be directly applied to diverse, unseen scenes without requiring model fine-tuning. However, it faces two significant challenges when applied to HSI. (i) The first challenge is the variation in spectral parameters across different hyperspectral sensors, including differences in spectral resolution, wavelength range, and the number of spectral channels [10]. For a P-T-based model like HyperSigma [54], although it is pre-trained using HSIs with fixed channels, the challenge can be addressed during its tuning stage by rebuilding the channel embedding layer to align the dimensions with the new data. But for a P-E-based model, the challenge of channel variation requires more careful consideration. (ii) The second challenge is to make prompt input tractable, ensuring a P-E based model can directly process unseen HSIs with least prior knowledge. Specifically, most existing P-E-based RSFMs rely on a frozen SAM [17], which segments remote sensing objects based on specialized models to generate prompts [4, 25]. As shown in Figure 1(b), each prompt typically generates a single mask and all the prompts for target objects are needed by existing P-E-based RSFMs, which largely reduced the model feasibility.

To address the above challenges, as shown in Figure 1 (c), we propose *a channel-adaptive, tuning-free hyperspectral foundation model, called HyperFree*, which can process any unseen HSIs directly in a P-E manner. Specifically, (i) Motivated by the word dictionary in natural language processing, which stores vector embeddings as values corresponding to words and phrases of varying lengths as keys [32], HyperFree introduces a learnable wavelength weight dictionary in the embedding layer, spanning the 400 to 2500 nm spectral range in 10 nm intervals. This allows for the dynamic encoding of unseen HSIs with varying spectral parameters into visual tokens, all with consistent dimensionality. (ii) To make the prompt of HyperFree more tractable, HyperFree is designed to generate multiple semantic-aware masks for one prompt as in Figure 1 (c), where HyperFree maps each prompt and single mask from image space into feature space and treats the feature distance as semantic-similarity. Prompt, mask and feature are interacted adaptively for different downstream task in tuning-free manner with only one prompt or zero shot.

Due to the high acquisition cost and label difficulty, current hyperspectral datasets only have a few images [45, 69], far from the requirements to train the promptable HyperFree. To conquer this problem, we developed the Hyper-Seg data engine, in addition to utilizing well-known multispectral datasets (e.g., fMoW [5] and SpaceNet [50, 51, 59]). This engine automatically generates a large-scale annotated HSI dataset, consisting of nearly 50,000 pairs of HSIs and their corresponding segmentation masks. The original data are sourced from AVIRIS airborne hyperspectral sensors, with a spatial size of 512×512 pixels and 224 spectral channels spanning from 400 to 2500 nm at a 10 nm spectral resolution. Unlike other RSFMs trained on unannotated satellite hyperspectral imagery with a 30m spatial resolution [54], the Hyper-Seg data engine features higher spatial resolution (0.6 to 5.0 m), along with annotations of 15.44 million masks. It is currently the largest known hyperspectral dataset with high spatial resolution.

To demonstrate the model capabilities, we evaluated HyperFree directly in a P-E manner on 11 datasets across 5 segmentation-related tasks, including multi-class classification [21], one-class classification [66], target detection [65], anomaly detection [47], and change detection [26]. Additionally, we verified HyperFree in a P-T manner on 14 datasets spanning 8 tasks, with additional tasks including hyperspectral denoising [27], hyperspectral unmixing [37], and hyperspectral object tracking [61]. The experimental results show that HyperFree achieves comparable accuracy to state-of-the-art models in a tuning-free manner (using just one prompt), and outperforms most models after tuning.

Our contributions can be summarized as:

- We propose the first tuning-free hyperspectral foundation model, which can process any hyperspectral image in different tasks with promptable or zero-shot manner.
- A weight dictionary that spans the full spectrum, enabling dynamic generation of the embedding layer dynamically according to input wavelengths.
- We propose to map both prompts and masks into feature space to identify multiple semantic-aware masks for one prompt, where different interaction workflows are designed for each downstream tasks.
- We built the Hyper-Seg data engine to train the HyperFree model and tested it on 11 datasets from 5 tasks in tuning-free manner, 14 datasets from 8 tasks in tuning manner as an extensive experiment.

## 2. Related Work

**Hyperspectral image processing.** There are diverse tasks to process such precise observation with different targets

[31]. (a) Classification: Assigning a specific class to each pixel [21]. (b) One class classification: Discriminating the pixels of target class with labeled binary map [33]. (c) Target detection. Discriminating the pixels of target class with labeled spectrum. (d) Anomaly detection. Discriminating the pixels spectrally deviating from the background [24]. (e) Change detection: Discriminating the changing pixels between two time-steps [26]. Since the output of the above tasks is all pixel-level segmentation maps, it is feasible to unify them using the proposed HyperFree model in tuning-free manner.

**Foundation models.** Powered by the transformer architectures, foundation models are characterized by large-scale property in model size and pre-training data, and the transferring ability for downstream tasks [60]. In remote sensing community, some foundation models for multi-spectral images have already been proposed such as SatMAE [7], SepctralEarth [1] and SpectralGPT [13], which trains large ViT-like models [9] with the MAE strategy [12] and massive open-source multi-spectral images (13 channels). In contrast, hyperspectral images have a high acquiring cost [56] and make the large-scale pre-training stage more difficult. A recent study broke the situation by building a large-scale hyperspectral dataset HyperGlobal-450K and trained a foundation model called HyperSigma with MAE strategy as well [54]. Our work differs from HyperSigma in two aspects: (i) HyperSigma accepts fixed number of image channels, while HyperFree can deal with any number of channels directly. (ii) HyperSigma needs to be fine-tuned for each downstream task. Differently, HyperFree has ability to process different tasks in fine-tuning free manner.

**Prompt engineering (P-E).** Different from the pre-training and then fine-tuning paradigm, P-E can utilize pre-trained foundation models to complete downstream tasks directly with prompts [34]. For pre-trained large language models, prompts are mostly hand-crafted text to guide the model generate response for specific task [34, 42]. In computer vison community, SAM follows the same spirit and designs diverse prompts (e.g., point and box) to achieve few-shot transferring performance [17]. Our work extends the visual prompt engineering in hyperspectral data with varying image channels and supports five tasks in tuning-free manner with designed prompt-mask-feature interaction.

# 3. Hyper-Seg Data Engine

Hyperspectral image processing always refers to various dense tasks and a large-scale segmentation dataset is needed to apply the P-E. However, most hyperspectral segmentation datasets only have a few images due to the high acquisition and label cost, which is quite different from the million-scale dataset with natural images such as SA-1B [17]. To tackle this, we built a data engine called Hyper-Seg to generate segmented masks automatically for spectral images and expand the data scale. Hyper-Seg engine firstly separates the spectral image into several groups with three channels according to the prior of key channels position. Each group is then segmented by the SAM-H model [17] and the results are combined by NMS operation to eliminate redundance and output the final segmentation map. Figure 2 shows the overall workflow, where the nine key channels are selected according to the expert knowledge as in Supp.1.1. Considering the high spatial resolution property of airborne platform, we collected 41946 hyperspectral images from AVIRIS airborne sensor and applied the Hyper-Seg Engine. Besides, we also processed two existing multispectral datasets (fMoW [5] and SpaceNet [50, 51, 59]) to increase the scale further. Compared to the traditional hyperspectral annotation methods which refer to high-resolution RGB imagery, our data engine can utilize more spectral information and obtain a high-precision segmentation mask without labor cost.

Table 1 shows the comparable statistical information of training datasets between existing RSFMs and proposed HyperFree. Most models only concentrate on the RGB and multispectral images with channel number 3$\sim$13. Although the recent model HyperSigma [54] has collected a large-scale hyperspectral images, the spatial resolution (30m/pixel) is largely lower than us (0.6$\sim$5m/pixel), and the dataset scale (considering image size and number together) is relatively small. With the designed Hyper-Seg data engine, constructed dataset is the only one with segmentation masks supporting the promptable training. More detailed statistical information of our generated dataset is given in Supp.1.2. The final generated dataset has nearly 150k images and 15.44milion masks. We name both the data engine and generated dataset as the Hyper-Seg for simplicity.

# 4. HyperFree Foundation Model

HyperFree has two customed components upon the meta-architecture of SAM including channel-adaptive embedding and prompt-mask-feature (PMF) interaction as in Figure 3. Channel-adaptive embedding constructs the embedding layer dynamically according to varying input wavelength, and outputs tokens in same representation space. After the promptable training stage, PMF interaction discriminates the semantic-aware masks for downstream tasks by mapping both prompt and mask into feature space and treating feature distance as semantic similarity.

## 4.1. Channel-adaptive Embedding

To make HyperFree process any hyperspectral image, channel-adaptive embedding layer is firstly designed to map input image with varying channels into tokens with fixed length, inspired by the tokenizer in NLP community [32].

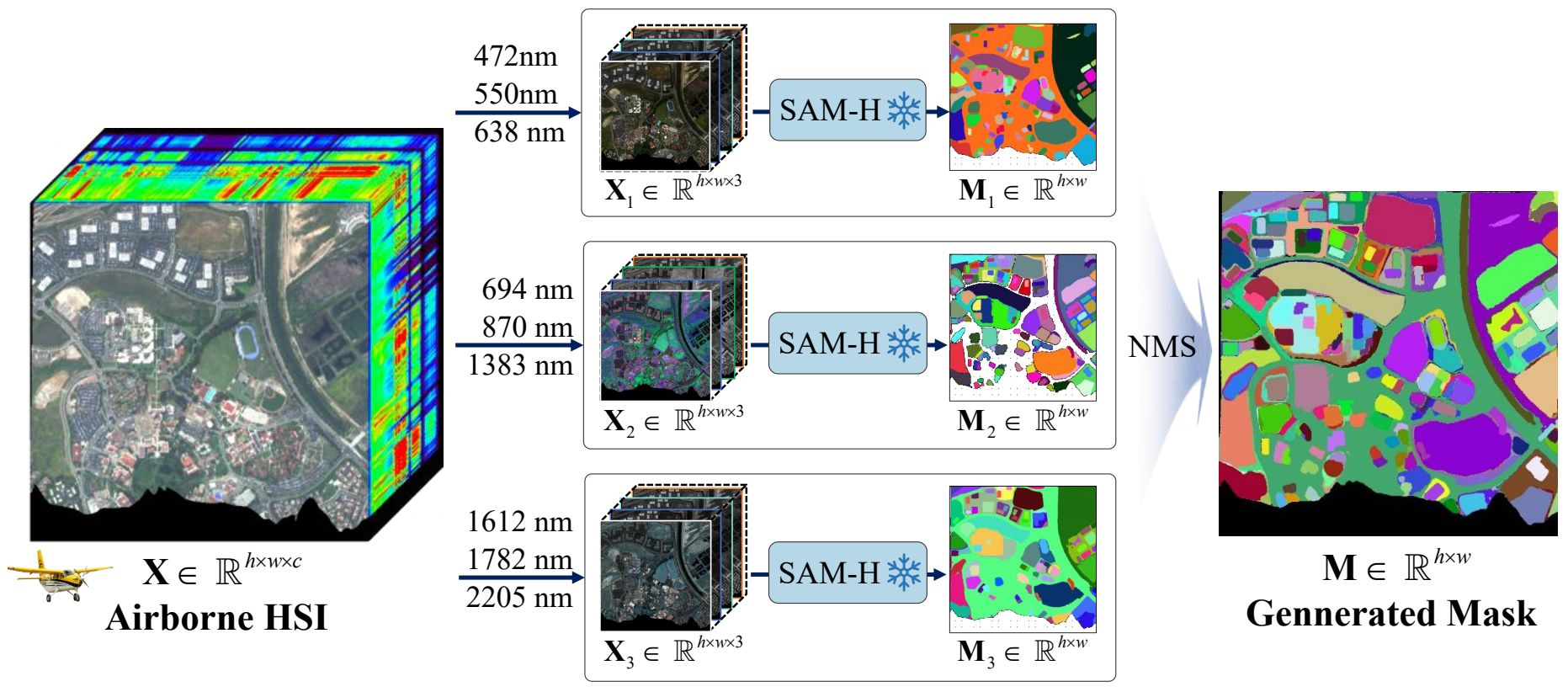

Figure 2. Hyper-Seg Data Engine automatically generates a large-scale dataset comprising nearly 41,900 pairs of airborne HSIs and their corresponding segmentation masks, enabling HyperFree to effectively learn discriminative spectral features.

Table 1. Comparison of training data used in current RSFMs. The primary features of the proposed HyperFree model include prompt engineering and the use of high-resolution annotated hyperspectral (HRS) imagery.

| Dataset | Foundation Model | Spatial Resolution (m) | Spectral Resolution (nm) | Channel Number | Image Size | Image Number | Main Modality | Training Paradigm |
|---|---|---|---|---|---|---|---|---|
| **RGB** | RVSA [53] | $0.5 \sim 153$ | \ | 3 | $110 \sim 31672$ | $1.00 \times 10^6$ | H/M/LSR-RGB | P-T |
| | RingMo [48] | $0.1 \sim 30$ | \ | 3 | 448 | $2.10 \times 10^6$ | H/MSR-RGB | P-T |
| | ScaleMAE [39] | $0.31 \sim 3.7$ | \ | 3 | \ | $3.64 \times 10^6$ | HSR-RGB | P-T |
| **Multispectral** | SatMAE [6] | 10 | $15 \sim 180$ | 13 | $\approx 45 \times 60$ | $7.13 \times 10^5$ | MSR-M | P-T |
| | SkySense [11] | 10/20 | $15 \sim 180$ | 10 | \ | \ | MSR-M | P-T |
| | SpectralGPT [14] | 10/20/60 | $15 \sim 180$ | 12 | $45 \sim 120$ | $1.47 \times 10^6$ | MSR-M | P-T |
| **Hyperspectral** | HyperSigma [54] | 30 | 5/6.5/10 | 150/175 | 64 | $4.47 \times 10^5$ | LSR-H | P-T |
| | HyperFree | $0.6 \sim 5.0$ | 10 | 224 | 512 | $4.19 \times 10^4$ | HSR-H | P-E |

* P-T: Pretraining and Tuning; P-E: Prompt Engineering.
* HSR: High Spatial Resolution; MSR: Medium Spatial Resolution; LSR: Low Spatial Resolution.
* RGB: Red-Green-Blue Imagery; M: Multispectral Imagery; H: Hyperspectral Imagery

Tokenizers rely on a pre-defined dictionary to break down a string of text into numerical tokens, where the dictionary has stored the numerical vectors of varied-length words or phrases. Vision transformers always use a convolutional layer to generate patch tokens and *what if we construct a layer weight dictionary treating each wavelength as a word*? Given the input meta data of wavelengths, we can just loop up the corresponding weights and construct the layer dynamically without re-training it.

The weight dictionary is built depending on the wavelengths to be processed and the expected token dimension. Since hyperspectral imaging mostly captures the spectral signals from 400nm to 2500nm and the 10nm is of high spectral resolution currently, 221 index channels are designed as the dictionary keys. Denote the output token as a vector of $\mathbb{R}^j$ corresponding to the patch size $p \times p$, each index channel stores a weight matrix of size $\mathbb{R}^{p\times p\times j}$, and the built dictionary $\beta$ has a size of $221 \times p \times p \times j$. Given the input hyperspectral image $\mathbb{R}^{h\times w\times n}$ with wavelengths $\mathbf{b} = [b_1, b_2, \ldots, b_n]$, the corresponding kernel weight can be dynamically generated by combining the searched weight $\mathbf{W}$ with size $n \times p \times p \times j$, where $n$ is flexible for each image. For conciseness, we use function $g$ to represent the mapping process from $\mathbf{b}$ to $\mathbf{W}$ as kernel weight, and function $f$ to represent the token generation step as in Equation (1).

$$f(\mathbf{X} \mid \mathbf{b}, \beta) = \mathrm{Conv}_{\mathbf{W}}(\mathbf{X}) \quad \text{where} \quad \mathbf{W} = g(\mathbf{b}, \beta) \qquad (1)$$

In the past few decades, many satellites have been launched equipped with spectral sensors, where the key channels are selected by professional teams and practical experience. To utilize these successful priors, we further extract visual tokens in two parallel branches with dynamically constructed layers: one to process key channels and the other to process the split cubes, as shown in Figure 3. The input image $\mathbf{X}$ is first divided into key channels $\mathbf{X_k}$ and the cubes $\mathbf{X_c}$ between them, where the key wavelengths are selected by referring to famous spectral sensors (detailed in Supp.2). We set two different dictionaries $\beta_\mathbf{k}$ and $\beta_\mathbf{c}$ to process $\mathbf{X_k}$ and $\mathbf{X_c}$ respectively, and the final output tokens $\mathbf{T} \in \mathbb{R}^{(h/p)\times(w/p)\times j}$ are computed as the element-wise sum, as shown in Equation (2).

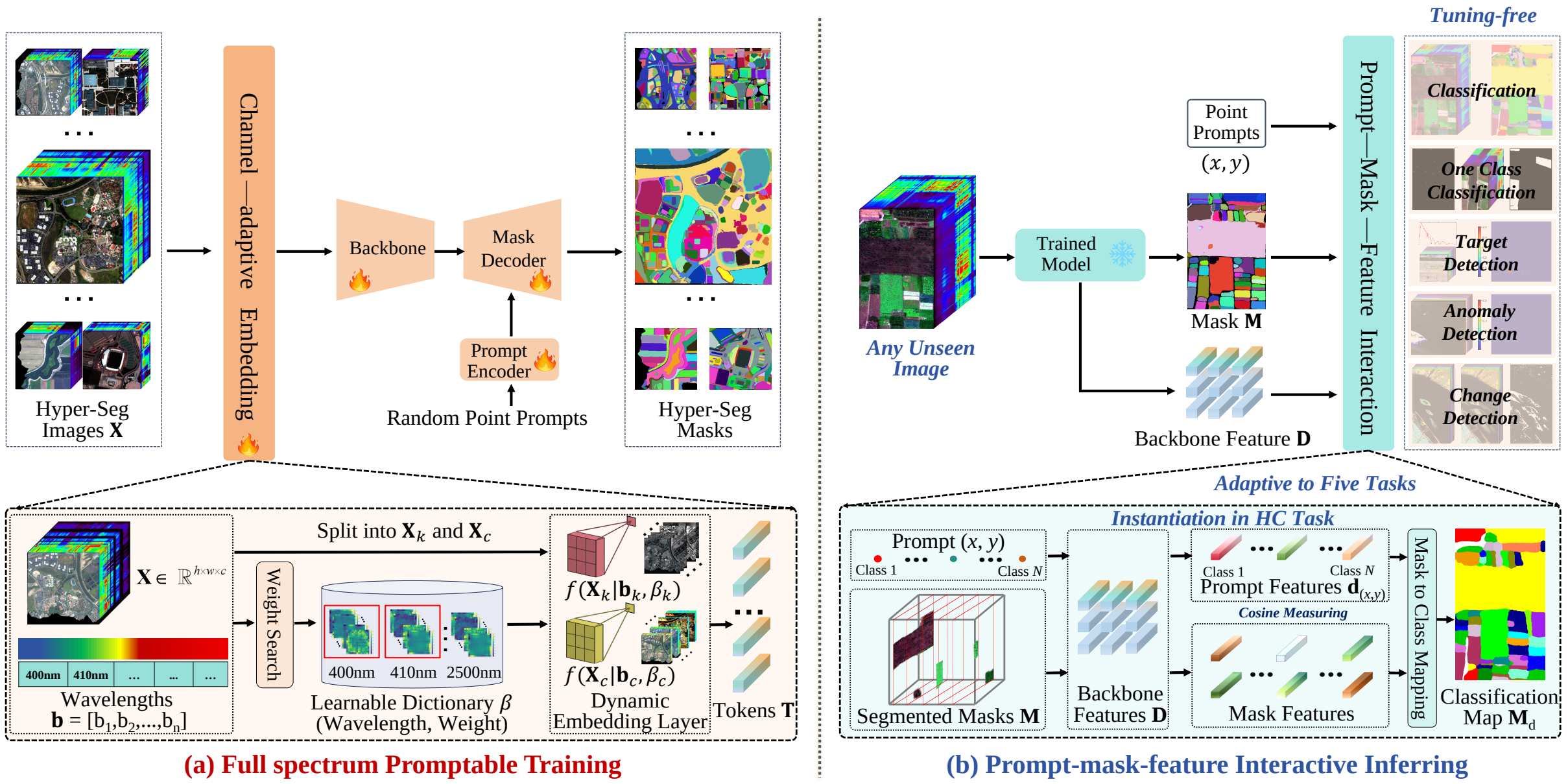

Figure 3. (a) Full spectrum Promptable Training. Based on the meta-architecture of promptable segmentation, channel-adaptive embedding layer is developed to process any-channel input, which relies on a learnable weight dictionary and generates the embedding layer dynamically according to the input wavelengths. With the constructed Hyper-Seg dataset, HyperFree is trained with randomly selected channels from full spectrum and point prompts in each iteration. (b) Prompt-mask-feature Interactive Inferring. Keep the trained HyperFree frozen, different PMF interaction workflows are designed to complete the downstream tasks in promptable or zero-shot manner, where we use feature distance to measure the semantic consistency. An interaction instantiation for classification task is given above as an example.

$$\mathbf{T} = f(\mathbf{X_k}|\mathbf{b_k}, \beta_\mathbf{k}) + f(\mathbf{X_c}|\mathbf{b_c}, \beta_\mathbf{c}) \quad (2)$$

## 4.2. Full-spectrum Promptable Training

The prompt engineering paradigm is adopted to complete downstream tasks directly with designed prompts. HyperFree is thus trained in a promptable manner on constructed large-scale Hyper-Seg dataset, where the prompts are randomly chosen at each iteration and the model is expected to segment correctly the corresponding masks. Since hyperspectral image processing tasks are almost conducted at the pixel level, we choose point prompts only at the training stage. Focal loss $L_f$ and Dice loss $L_d$ are weighted as [17] to supervise the training as Equation (3).

$$L = 20L_f + L_d \quad (3)$$

To train the channel-adaptive embedding layer effectively, we select the subset of channels randomly from the full-spectrum range and process it with corresponding weights only in $\beta_\mathbf{k}$ and $\beta_\mathbf{c}$, which helps the learned dictionary being wavelength-aware and encode the varied channel composition into a unified token space.

## 4.3. Prompt-mask-feature (PMF) Interactive Inferring

After training process, the model can process unseen hyperspectral image with varied channels and segment the corresponding mask given any point prompt. However, many downstream tasks need to segment out all the masks of same semantic category rather than a single mask such as hyperspectral classification and target detection. We tackle this problem by *connecting the prompt, mask and feature together in feature space*, where prompt tells the model an object reference in task, mask outputs accurate location information, and feature reflects semantic information. In unified feature space, the cosine distance of features is used to measure the semantic similarity.

Before diving into the specific workflows, we introduce two basic interaction modes with a simple situation. Set one point $(x, y)$ as input prompt for image $\mathbf{X}$, the corresponding feature cube $\mathbf{D}$ and all the generated masks $\mathbf{M} \in \mathbb{R}^{h\times w\times k}$ from the trained model, where $\mathbf{M}$ has $k$ binary maps and each represents a single mask. The first interaction mode maps the prompt $(x, y)$ into a feature vector $\mathbf{d}_{(x,y)} \in \mathbb{R}^j$ as in Equation (4), where the corresponding valid and smallest mask in $\mathbf{M}$ is used to localize the features in $\mathbf{D}$, and the mean feature is treated as $\mathbf{d}_{(x,y)}$ for the given prompt. This interaction transforms the prompt from image space to feature space, and makes it possible to explore the semantic similarity.

$$\text{Mode 1: } (x, y)|\mathbf{M}, \mathbf{D} \rightarrow \mathbf{d}_{(x,y)} \tag{4}$$

The second interaction mode selects multiple the masks in $\mathbf{M}$, which have similar semantic information with $\mathbf{d}_{(x,y)}$ as in Equation (5). The selected masks are represented as $\mathbf{M}_d$ and $\mathbf{M}_d \subseteq \mathbf{M}$. Specifically, we compute a feature representation for each single mask by localizing in $\mathbf{D}$ and computing the average, and the cosine similarity between them and $\mathbf{d}_{(x,y)}$ are treated as the semantic similarity. With some hand-designed thresholds $\tau$, $\mathbf{M}_d$ can be finally segmented. $\tau$ represents different meanings according to the specific task.

$$\text{Mode 2: } (\mathbf{d}_{(x,y)}|\mathbf{M}, \mathbf{D}, \tau) \rightarrow \mathbf{M}_d \tag{5}$$

By using both interaction modes adaptively, five hyperspectral tasks can be completed with the trained promptable model directly. (a) **Hyperspectral classification (HC)**. Each category needs at least one prompt, and uses Mode 1 and Mode 2 in turn for each category. For a common closed-set setting, $\tau$ is not necessary since we can use the minimum feature distance between each mask representation and all the prompt representations to decide its category. Figure 3 (b) shows an exemplified workflow. (b) **Hyperspectral one-class classification (HOCC)**. HOCC is similar to HC, but the difference lies in that HOCC only needs prompts of the target category, and its $\tau$ is instantiated as a class prior parameter to control the mask ratio. (c) **Hyperspectral target detection (HTD)**. Compared to HOCC, HTD accepts the target spectra and it needs to be changed to a prompt by selecting the pixel with the smallest cosine distance. $\tau$ is instantiated as a distance threshold to convert the density map to a binary map. (d) **Hyperspectral anomaly detection (HAD)**. HAD is an unsupervised task and there is no $\mathbf{d}_{(x,y)}$ for HAD. Since HAD is very similar to unsupervised target detection, we use $\mathbf{M}$ only and filter out large objects according to mask area to output the final anomaly map. (e) **Hyperspectral change detection (HCD)**. Different from the prior tasks, HCD accepts two registered and temporal images to identify the changing locations. The HCD task utilizes both interaction modes for the bi-temporal images. Mode 1 processes the first image to get representation $\mathbf{d}_{(x,y)}$ for each mask, which is input to Mode 2 and compared with the corresponding mask representation with $\mathbf{M}$ and $\mathbf{D}$ of the second image. Once the temporal feature distance exceeds the preset $\tau$, a change is considered to have happened. The interaction workflow is summarized in Table 2.

Table 2. Instantiation summary of prompts, meaning of $\tau$, and interaction modes when HyperFree performs different tasks.

| Task | Prompt Type | Hyperparameter $\tau$ | Interaction Mode |
|---|---|---|---|
| HC | One Point Per Categories | No $\tau$ | Mode 1, Mode 2 |
| HOCC | Target Category Point | Class Prior | Mode 1, Mode 2 |
| HTD | Target Spectra | Similarity Threshold | Mode 1, Mode 2 |
| HAD | No Prompt | Masks Area Threshold | Mode 1 |
| HCD | No Prompt | Similarity Threshold | Mode 1, Mode 2 |

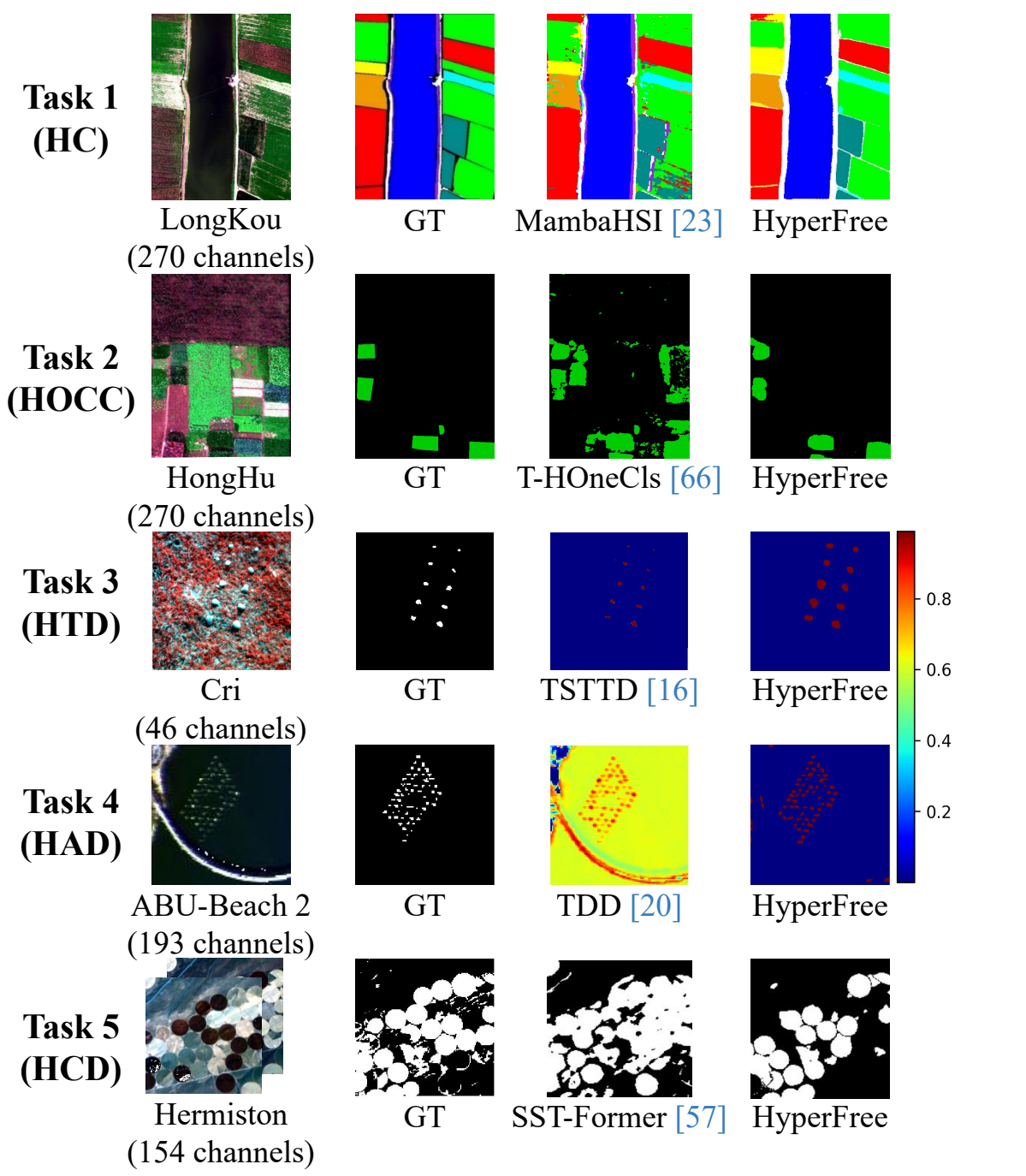

Figure 4. Exemplified comparison results for five tuning-free tasks.

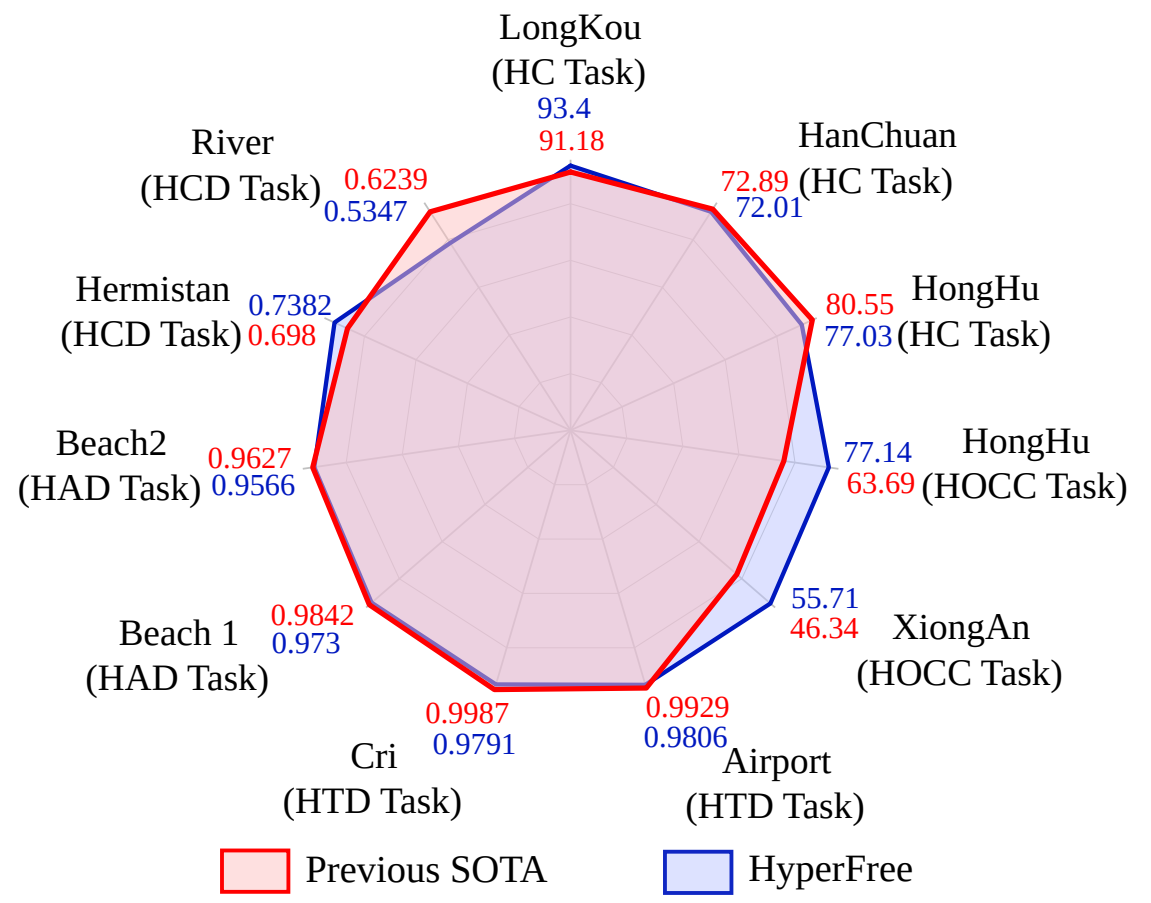

Figure 5. In tuning-free manner, HyperFree (1 prompt) can achieve comparable results with specialized models (5 shots) on 5 tasks and 11 datasets.

## 5. Results

### 5.1. Experimental Setup

We evaluate the proposed HyperFree on five downstream hyperspectral tasks including HC, HOCC, HTD, HAD and HCD tasks as in Section 4.3. A total of 11 datasets were used in the evaluation (detailed in Supp.3), and each task had six comparison methods, including both classical and state-of-the-art ones. Since there is no method that can directly complete all five tasks, all comparative methods are tested after being trained on each dataset with five shots per class, while the parameters of HyperFree are fixed for all datasets with 1 prompt or zero-shot as in Table 2. We use ViT-base as backbone to conduct the experiments.

Table 3. Quantitative comparison results on five hyperspectral tasks, where blue numbers indicates the metric ranking of HyperFree. Comparison methods were all trained on each dataset while HyperFree processed the datasets in tuning-free manner.

| HC Task | | | | | | | | | | | | | | |
|---|---|---|---|---|---|---|---|---|---|---|---|---|---|---|
| | LongKou (270 channels) | | | | | | | HanChuan (274 channels) | | | | | | |
| Metrics | SVM [30] (5 shot) | HybridSN [41] (5 shot) | FCN [52] (5 shot) | FPGA [68] (5 shot) | SSFTT [46] (5 shot) | MambaHSI [23] (5 shot) | HyperFree (1 prompt) | SVM [30] (5 shot) | HybridSN [41] (5 shot) | FCN [52] (5 shot) | FPGA [68] (5 shot) | SSFTT [46] (5 shot) | MambaHSI [23] (5 shot) | HyperFree (1 prompt) |
| OA | 82.77 | 48.78 | 86.67 | 91.18 | 89.66 | 92.65 | 93.39(1st) | 52.68 | 47.75 | 55.55 | 71.47 | 64.86 | 73.33 | 75.47(1st) |
| AA | 74.02 | 61.37 | 85.60 | 88.35 | 87.96 | 92.57 | 88.03(3rd) | 47.76 | 46.17 | 59.72 | 72.09 | 61.22 | 69.33 | 62.35(2nd) |
| KA | 78.04 | 35.72 | 82.30 | 88.66 | 87.95 | 90.00 | 91.41(1st) | 46.85 | 41.31 | 50.18 | 67.58 | 59.65 | 69.10 | 71.56(1st) |
| HOCC Task | | | | | | | | | | | | | | |
| | HongHu (270 channels) | | | | | | | XiongAn (256 channels) | | | | | | |
| Metrics | OCSVM [43] (5 shot) | nnPU [18](5 shot) | BSVM [33](5 shot) | PAN [15](5 shot) | OC Loss [67](5 shot) | T-HOneCls [66](5 shot) | HyperFree (1 prompt) | OCSVM [43] (5 shot) | nnPU [18](5 shot) | BSVM [33](5 shot) | PAN [15](5 shot) | OC Loss [67](5 shot) | T-HOneCls [66](5 shot) | HyperFree (1 prompt) |
| $F_1$ | 26.33 | 19.13 | 34.82 | 63.69 | 54.73 | 55.97 | 72.52(1st) | 18.31 | 1.76 | 26.30 | 46.34 | 43.08 | 41.34 | 52.50(1st) |
| P | 56.43 | 19.72 | 50.77 | 80.00 | 58.27 | 46.52 | 68.61(2nd) | 39.83 | 2.85 | 23.82 | 47.13 | 47.50 | 32.87 | 68.21(1st) |
| R | 24.02 | 18.58 | 45.29 | 64.27 | 54.34 | 92.35 | 78.06(2nd) | 16.08 | 1.98 | 57.83 | 53.32 | 47.61 | 60.38 | 65.03(1st) |
| HTD Task | | | | | | | | | | | | | | |
| | Airport (205 channels) | | | | | | | Cri (46 channels) | | | | | | |
| Metrics | ACE [19](1 shot) | CEM [2](1 shot) | GLRT [28](1 shot) | MF [29](1 shot) | HTD-IRN [44](1 shot) | TSTTD [16](1 shot) | HyperFree (1 prompt) | ACE [19](1 shot) | CEM [2](1 shot) | GLRT [28](1 shot) | MF [29](1 shot) | HTD-IRN [44](1 shot) | TSTTD [16](1 shot) | HyperFree (1 prompt) |
| $AUC_{(D,F)}$ | 0.9794 | 0.9603 | 0.9801 | 0.9916 | 0.9745 | 0.9929 | 0.9806(3rd) | 0.9735 | 0.9893 | 0.9737 | 0.9891 | 0.9975 | 0.9987 | 0.9791(5th) |
| $AUC_{ODP}$ | 1.5853 | 1.2829 | 1.5798 | 1.6968 | 1.4484 | 1.6592 | 1.4667(6th) | 1.2015 | 1.4506 | 1.2000 | 1.4575 | 1.3995 | 1.6103 | 1.4670(2nd) |
| HAD Task | | | | | | | | | | | | | | |
| | Beach1 (188 channels) | | | | | | | Beach2 (193 channels) | | | | | | |
| Metrics | RXD [40] | CRD [22] | ADLR [36] | LRASR [63] | Auto-AD [55] | TDD [20] | HyperFree (zero shot) | RXD [40] | CRD [22] | ADLR [36] | LRASR [63] | Auto-AD [55] | TDD [20] | HyperFree (zero shot) |
| $AUC_{(D,F)}$ | 0.9815 | 0.9471 | 0.4515 | 0.7461 | 0.9574 | 0.9842 | 0.9730(3rd) | 0.9090 | 0.8544 | 0.7976 | 0.8225 | 0.9485 | 0.9627 | 0.9566(2nd) |
| $AUC_{ODP}$ | 1.2557 | 0.9785 | 0.5610 | 0.8526 | 1.1273 | 1.1383 | 1.9191(1st) | 1.0177 | 0.8670 | 0.9064 | 0.8280 | 1.0097 | 1.1688 | 1.8697(1st) |
| HCD Task | | | | | | | | | | | | | | |
| | River (154 channels) | | | | | | | Hermiston (198 channels) | | | | | | |
| Metrics | FC-EF [8](5 shot) | FC-Sc [8](5 shot) | FC-Sd [8](5 shot) | ML-EDAN [35](5 shot) | BIT [3](5 shot) | SST-Former [57](5 shot) | HyperFree (zero shot) | FC-EF [8](5 shot) | FC-Sc [8](5 shot) | FC-Sd [8](5 shot) | ML-EDAN [35](5 shot) | BIT [3](5 shot) | SST-Former [57](5 shot) | HyperFree (zero shot) |
| IoU | 0.4168 | 0.4522 | 0.4534 | 0.3915 | 0.2126 | 0.4096 | 0.3649(6th) | 0.3729 | 0.3776 | 0.4873 | 0.3252 | 0.5257 | 0.5361 | 0.5851(1st) |
| $F_1$ | 0.5884 | 0.6228 | 0.6239 | 0.5628 | 0.3507 | 0.5812 | 0.5347(6th) | 0.5432 | 0.5482 | 0.65.52 | 0.4908 | 0.6891 | 0.6980 | 0.7382(1st) |

Table 4. Compared results with SAM using the identical PMF interaction workflows.

| | HC | | | HOCC | | | HTD | | HAD | | HCD | |
|---|---|---|---|---|---|---|---|---|---|---|---|---|
| | OA | AA | KA | $F_1$ | Precision | Recall | $AUC_{(D,F)}$ | $AUC_{ODP}$ | $AUC_{(D,F)}$ | $AUC_{ODP}$ | $F_1$ | IoU |
| SAM+PMF Interaction | 66.62 | 72.99 | 59.41 | 34.13 | 30.35 | 39.47 | 0.7979 | 1.1014 | 0.9615 | 1.8844 | 0.5996 | 0.4282 |
| HyperFree | 93.40 | 85.66 | 91.42 | 72.52 | 68.61 | 78.06 | 0.9791 | 1.4670 | 0.9566 | 1.8697 | 0.7382 | 0.5851 |

HyperFree was trained on the constructed Hyper-Seg dataset using 8 A100 GPUs for approximately 10 hours, which was based on the meta-architecture and parameters of SAM. We use AdamW optimizer with initial learning rate 8e-5 and batch size 16. To help learning the channel-adaptive embedding, we randomly select the image channels at each iteration and match it with corresponding wavelengths. 16 masks and 1$\sim$2 points per mask are randomly selected for each input image to train the promptable segmentation.

### 5.2. Main Results

Most tuning-free quantitative results are reported in Table 3 (1 dataset in supplement due to the space limit) and Figure 5. All the comparative methods are trained with 5 shot for each class and HyperFree processes each dataset directly without tuning. For HC, HOCC and HTD, only one point prompt is provided for each class. HAD and HCD does not need any prompt and deal with images in zero-shot manner. Table 3 shows than HyperFree can surpass most of the specialized models with frozen parameters. On both HC and HCD tasks, HyperFree even outperformed previous SOTA results, including an improvement of about 12 points in $F_1$ on HC and about 0.5 points in IoU on HCD task.

We performed qualitative comparison on five exemplified images in Figure 4, where one SOTA model have been visualized for each task comparison. Obviously, the result maps of HyperFree are more complete and smoother than the ones of specialized models. In HC, HOCC, HTD and HAD tasks, HyperFree can generate basically the same result as GT, which are difficulty for existing models.

Table 5. Ablation results on the usage of the learned weight dictionary.

| Correct Wavelengths | $\beta_{\mathbf{k}}$ | $\beta_{\mathbf{c}}$ | HC | | | HOCC | | | HTD | | HAD | | HCD | |
|---|---|---|---|---|---|---|---|---|---|---|---|---|---|---|
| | | | OA | AA | KA | $F_1$ | Precision | Recall | $AUC_{(D,F)}$ | $AUC_{ODP}$ | $AUC_{(D,F)}$ | $AUC_{ODP}$ | $F_1$ | IoU |
| ✓ | ✓ | × | 73.18 | 64.23 | 65.85 | 65.62 | 47.30 | 64.56 | 0.9767 | 1.4644 | 0.9488 | 1.8465 | 0.5793 | 0.4078 |
| ✓ | × | ✓ | 66.63 | 71.52 | 59.44 | 63.62 | 58.61 | 78.44 | 0.9768 | 1.4640 | 0.9589 | 1.8768 | 0.7022 | 0.5411 |
| × | ✓ | ✓ | 75.25 | 38.98 | 66.44 | 57.30 | 64.09 | 54.83 | 0.9595 | 1.4273 | 0.9557 | 1.8678 | 0.6832 | 0.5188 |
| ✓ | ✓ | ✓ | 93.40 | 85.66 | 91.42 | 72.52 | 68.61 | 78.06 | 0.9791 | 1.4670 | 0.9566 | 1.8697 | 0.7382 | 0.5851 |

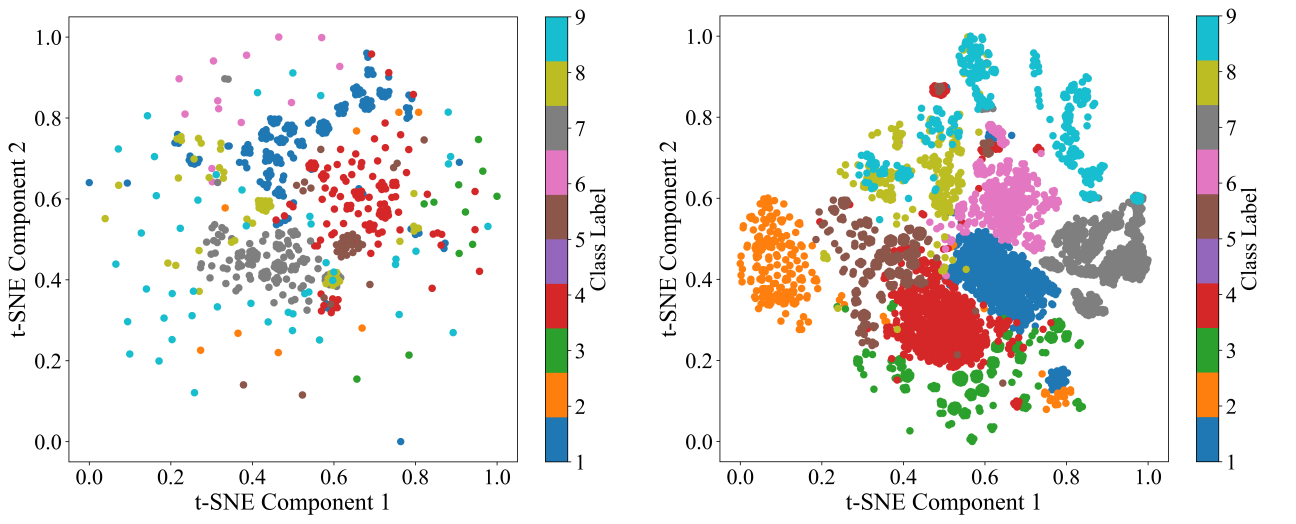

(a) HyperSigma features (b) HyperFree features

Figure 6. t-SNE visualizations of HyperSigma [54] and HyperFree features on WHU-Hi Longkou dataset. We randomly select 1000 pixels for each class.

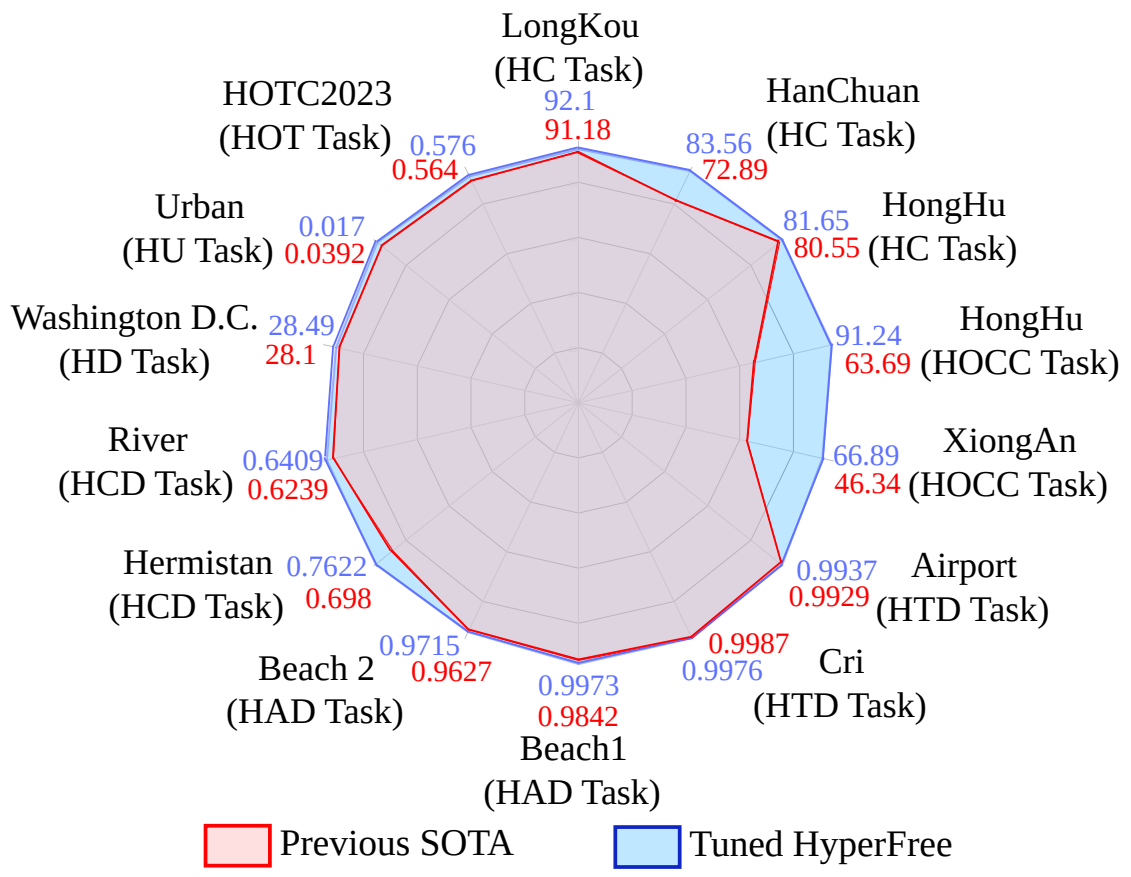

Figure 7. In tuning manner, HyperFree can achieve overall better results with specialized models on 8 tasks and 14 datasets.

### 5.3. Analysis Studies

**HyperFree vs SAM**. Proposed prompt-mask-feature interaction strategy adapts the pre-trained HyperFree to various downstream tasks. We applied the same strategy to SAM to verify the effectiveness of training on hyperspectral images and reported the results in Table 4. The metrics on most tasks have a decline of about 10∼30 points, implying a failed processing. We realize it is not a fair comparison and only use it to reflect the difference of hyperspectral data processing.

**HyperFree vs HyperSigma**. HyperSigma is the only open-sourced hyperspectral foundation model [54, 62], which adopts the pre-training and tuning paradigm. Thus, we cannot compare HyperFree with it in tuning-free setting directly and compare the t-SNE visualization in Figure 6(a) and Figure 6(b). In contrast, proposed HyperFree has a more compact intra-class distance obviously, supporting the tuning-free workflow of PMF interaction.

**Full-Spectrum Weight Dictionary**. To verify stored weights are wavelength-aware, we conducted related ablation studies in Table 5. In HC, HOCC and HCD tasks, a drop in accuracy of about 5∼40 points can be observed if we shuffled the correct wavelength order randomly, showing the wavelength-aware property of learned dictionary. Besides, we proved the built two encoding branches with dictionaries $\beta_{\mathbf{k}}$ and $\beta_{\mathbf{c}}$ can benifit each other effectively.

**Tuning HyperFree**. As an extensive experiment, we have also tested the tuning performance of HyperFree on 8 tasks and 14 datasets in Figure 7, which further included hyperspectral denosing (HD) [27], object tracking (HOT) [61] and unmxing (HU) tasks [37]. The tuning HyperFree has surpassed the previous SOTA models on nearly all tasks.

**More analysis**. For more analytical experiments, please see the supplementary material. (i) Supp.4.1: Complete qualitative comparison of 11 datasets in tuning-free manner. (ii) Supp.4.2: Complete quantitative and qualitative comparison of 14 datasets in tuning manner. (iii) Supp.4.3: Sensitivity analysis about prompt number, prompt location and the hyperparameter $\tau$.

## 6. Conclusion

Hyperspectral images can provide rich spectral information and are widely used in resource monitoring and defense fields. However, its unique properties increase the difficulty of foundation model development such as varied-length channels, high sensitivity to imaging conditions and the acquirement difficulty, making the mainstream P-T being low cost-effectiveness due to the image-by-image tuning. Our work tackles this problem by converting to adapt the visual P-E [17]. Differently, we design a learnable weight dictionary covering full-spectrum to generate the channel-adaptive embeddings and PMF interaction to generate semantic-aware masks for different tasks. We hope HyperFree can speed up the era of "GPT moment" for hyperspectral intelligent processing.

## Acknowledgement

This work was supported by the National Natural Science Foundation of China under Grant No. 42325105 and Grant 424B2010.

# HyperFree: A Channel-adaptive and Tuning-free Foundation Model for Hyperspectral Remote Sensing Imagery

Supplementary Material

## 1. Hyper-Seg Data Engine

### 1.1. Wavelength Selection

To utilize the spectral information rather than only the RGB channels, Hyper-Seg engine separates the original image into 3 groups with different wavelength combination, where we refer to the famous Landsat-8 satellite in Table 1. The selected wavelengths represents the valuable practical experience and cover the overall range from $0.4 \sim 2.5\,\mu$m.

Table 1. Statistical results of classical multispectral satellites about the selected wavelengths, supporting the wavelength selection of Hyper-Seg data engine and the weight dictionary $\boldsymbol{\beta_k}$.

| Satellite | Central Wavelengths(nm) |
|---|---|
| Landsat-7 | 482.5, 565, 660, 825, 2220, 1650, 11450 |
| Landsat-8 | 443, 482.5, 562.5, 655, 865, 1610, 1375, 2200, 10895, 12005 |
| Sentinel-2A/2B | 443, 490, 560, 665, 705, 740, 783, 842, 865, 945, 1375, 1610, 2190 |
| WorldView-2/3 | 425, 480, 545, 605, 660, 725, 832.5, 950 |
| ZY1-02D/E | 486.5, 564.5, 662.5, 835.5, 434, 612, 730, 959 |
| ZY-3 | 485, 555, 660, 830 |
| RapidEye | 455, 555, 655, 710, 805 |
| PlanetScope | 485, 545, 630, 820 |
| GeoEye-1 | 480, 545, 672.5, 850 |
| SPOT-6/7 | 485, 560, 655, 825 |
| Pleiades-1A/B | 490, 560, 650, 840 |
| IRS-P6 | 555, 640, 815, 1625 |
| KOMPSAT-2/3/4 | 485, 560, 660, 830 |
| GF-1/2 | 485, 555, 660, 830 |
| GF-4 | 485, 560, 660, 830, 3800 |
| GF-6 | 485, 555, 660, 830, 720, 750, 425, 610 |

### 1.2. Statistical Information

Figure 1 reports the statistical information of constructed Hyper-Seg dataset. With the non-maximum suppression (NMS) operation, the number of final combined masks is approximately 2 to 3 times the number of masks for each group separately as in Figure 1 (a), indicating that Spectral-Seg can utilize the spectral information effectively. From Figure 1 (b) and Figure 1 (c), it can be observed that the number density of generated masks in the three source datasets is roughly equivalent and the small masks dominate the dataset, increasing the segmentation difficulty.

## 2. Selection of Key Channels in Weight Dictionary

In proposed channel-adaptive embedding layer, we design a sperate branch for processing key channels, which are set according to the successful prior of launched satellites in Table 1. Each wavelength in Table 1 is selected by expert knowledge. To merge the wavelengths that almost overlap between different satellites, we sort all the wavelengths first, take the average of every two adjacent wavelengths with interval less than 10nm (common spectral resolution) and substitute them. Combining with the longest wavelength 2500nm, a total of 85 wavelengths were selected to build the learnable dictionary $\boldsymbol{\beta_k}$.

## 3. Overview of Experimental Datasets

All the used public datasets are summarized in Table 2, which have different channel numbers and spectral ranges. We have tested both the tuning-free manner and tunning manner in five tasks including HC, HOCC, HTD, HAD and HCD. Due to the different output formats, only tuning manner is applied on HD, HU and HOT tasks.

Table 2. Summary of used public datasets on the eight tasks.

| Tasks | Datasets | Number of Channels | Spectral Range (nm) |
|---|---|---|---|
| HC | LongKou [55] | 270 | 400 $\sim$ 1000 |
| | HanChuan [55] | 274 | 400 $\sim$ 1000 |
| | HongHu [55] | 270 | 400 $\sim$ 1000 |
| HOCC | HongHu [55] | 270 | 400 $\sim$ 1000 |
| | XiongAn [50] | 256 | 390 $\sim$ 1000 |
| HTD | Airport [2] | 205 | 400 $\sim$ 2500 |
| | Cri [51] | 46 | 650 $\sim$ 1100 |
| HAD | Beach-1 [2] | 188 | 430 $\sim$ 860 |
| | Beach-2 [2] | 193 | 430 $\sim$ 860 |
| HCD | Hermiston [12] | 154 | \ |
| | River [12] | 198 | 400 $\sim$ 2500 |
| HD | Washington D.C. [3] | 191 | 400 $\sim$ 2400 |
| HU | Urban [15] | 162 | 400 $\sim$ 2500 |
| HOT | HOTC 2023 [1] | 56 | 460 $\sim$ 960 |

## 4. Additional Experiments

### 4.1. Qualitative Results in Tuning-free Manner

Complete visualization results are shown for five tasks. (a) Figure 3, 4 and 5 for HC task. (b) Figure 6 and 7 for HOCC task . (c) Figure 9 and 8 for HTD task. (d) Figure 10 and 11 for HAD task. (e) Figure 12 and 13 for HCD task. With the powerful segmentation ability and PMF interaction, HyperFree can achieve the best visualization performance without tuning compared to the specialized models training with 5 shots.

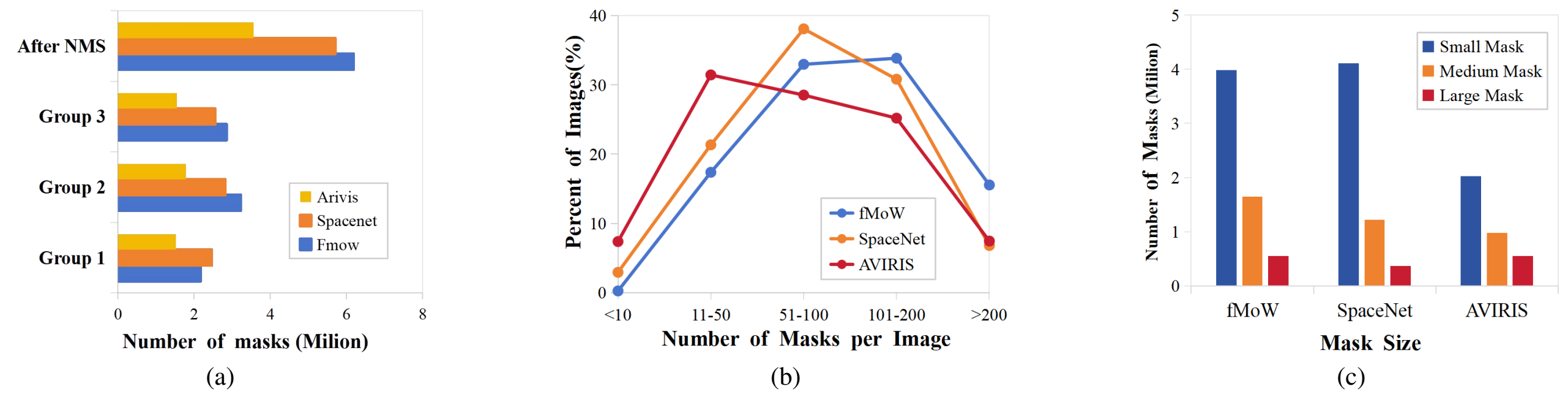

Figure 1. Some statistical information about the built large-scale Hyper-Seg dataset.

## 4.2. Quantitative and Qualitative Results With Tuning

We have also tested the further tuning performance of HyperFree as an extensive experiment. The tuned version is denoted as HyperFree* for simplicity. The quantitative results are reported in Table 4 ∼ Table 11 for HC, HOCC, HTD, HAD, HCD, HD, HU and HOT tasks, respectively. For the five tasks supporting the tuning-free manner, the qualitative results of HyperFree* are put together with results in Section 4.1. The qualitative results of HD, HU and HOT tasks are shown in Figure 14, 15 and 16, respectively. After tuning, HyperFree has achieved the best performance in most datasets and tasks. Since HyperFree is proposed mainly for tuning-free manner, we use the full-tuning setting directly without using any advanced tuning methods.

Table 3. Execution time comparison with deep models on five tuning-free tasks.

| | | | |
|---|---|---|---|
| HC | SSFTT [41] | MambaHSI [22] | HyperFree |
| | 197.08s | 429.86s | 7.85s(1st) |
| HOCC | OC Loss [53] | T-HOneCls [52] | HyperFree |
| | 244.46s | 344.48s | 21.50s(1st) |
| HTD | HTD-IRN [40] | TSTTD [14] | HyperFree |
| | 25.05s | 378.40s | 9.85s(1st) |
| HAD | Auto-AD [43] | TDD [18] | HyperFree |
| | 78.81s | 28.31s | 11.72s(1st) |
| HCD | BIT [5] | SST-Former [45] | HyperFree |
| | 142.23s | 116.05s | 15.90s(1st) |

## 4.3. Sensitivity Analysis

**Prompt Number**. In the five tuning-free tasks, HC and HOCC need prompts of each category to generate the semantic-aware results. We explored the relationship between model performance and the number of prompts as in Figure 2. The mean and std of metrics are calculated for each prompt number with 10 repeat experiments. We found HyperFree is mostly insensitive to the prompt number in both tasks and one prompt is good enough.

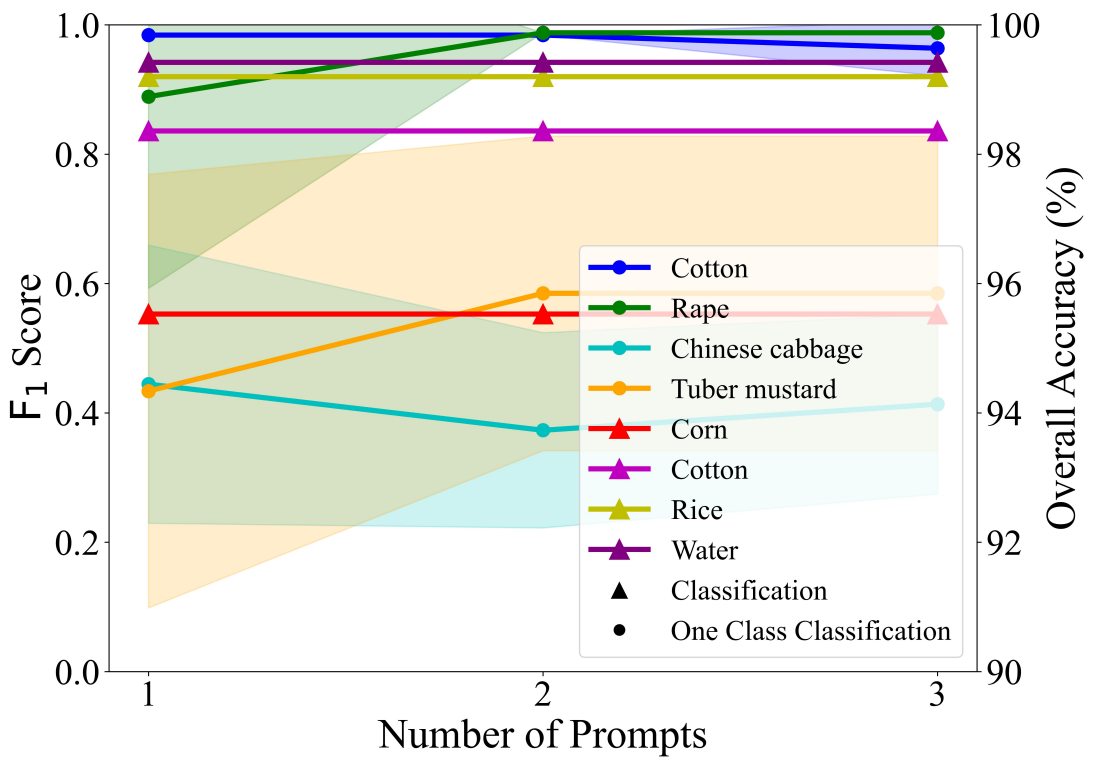

Figure 2. Sensitivity analysis of the prompt number on the model performance (HC and HOCC tasks).

**Hyperparameter** $\tau$. HyperFree completes five tasks directly with the PMF interaction, where the two interaction modes are used adaptively with the hyperparameter $\tau$. To explore its sensitivity, we have varied it and reported the corresponding results in Figure 17. HC task is not included since it does not need any $\tau$. Most tasks show a certain but acceptable sensitivity to $\tau$, where the HTD and HCD tasks exhibit more variation. Despite this, the fluctuation range of the metrics remains within an acceptable range of 0.1.

## 4.4. Execution Efficiency Comparison Experiments

Without the tuning process, HyperFree can reduce the processing time by 1 ∼ 2 orders of magnitude compared to other deep models as in Table 3.

Table 4. Quantitative comparison results on HC task in tuning manner, where HyperFree* represents the tuning version and blue numbers indicate the metric ranking.

| Dataset | Metric | SVM [28] (5 shot) | HybridSN [38] (5 shot) | FullyContNet [42] (5 shot) | FPGA [54] (5 shot) | SSFTT [41] (5 shot) | MambaHSI [22] (5 shot) | HyperFree* (5 shot) |
|---|---|---|---|---|---|---|---|---|
| LongKou [55] | OA | 82.77 | 48.78 | 86.67 | 91.18 | 89.66 | 92.65 | 92.10(2nd) |
| | AA | 74.02 | 61.37 | 85.6 | 88.35 | 87.96 | 92.57 | 92.71(1st) |
| | KA | 78.04 | 35.72 | 82.3 | 88.66 | 87.95 | 90.00 | 89.85(2nd) |
| HanChuan [55] | OA | 52.68 | 47.75 | 55.55 | 71.47 | 64.86 | 73.33 | 83.56(1st) |
| | AA | 47.76 | 46.17 | 59.72 | 72.09 | 61.22 | 69.33 | 82.21(1st) |
| | KA | 46.85 | 41.31 | 50.18 | 67.58 | 59.65 | 69.1 | 81.03(1st) |
| HongHu [55] | OA | 52.89 | 31.22 | 55.84 | 80.55 | 64.31 | 78.96 | 81.65(1st) |
| | AA | 45.97 | 34.14 | 67.25 | 75.12 | 64.53 | 76.02 | 85.56(1st) |
| | KA | 45.47 | 24.51 | 49.67 | 75.74 | 57.79 | 73.29 | 77.58(1st) |

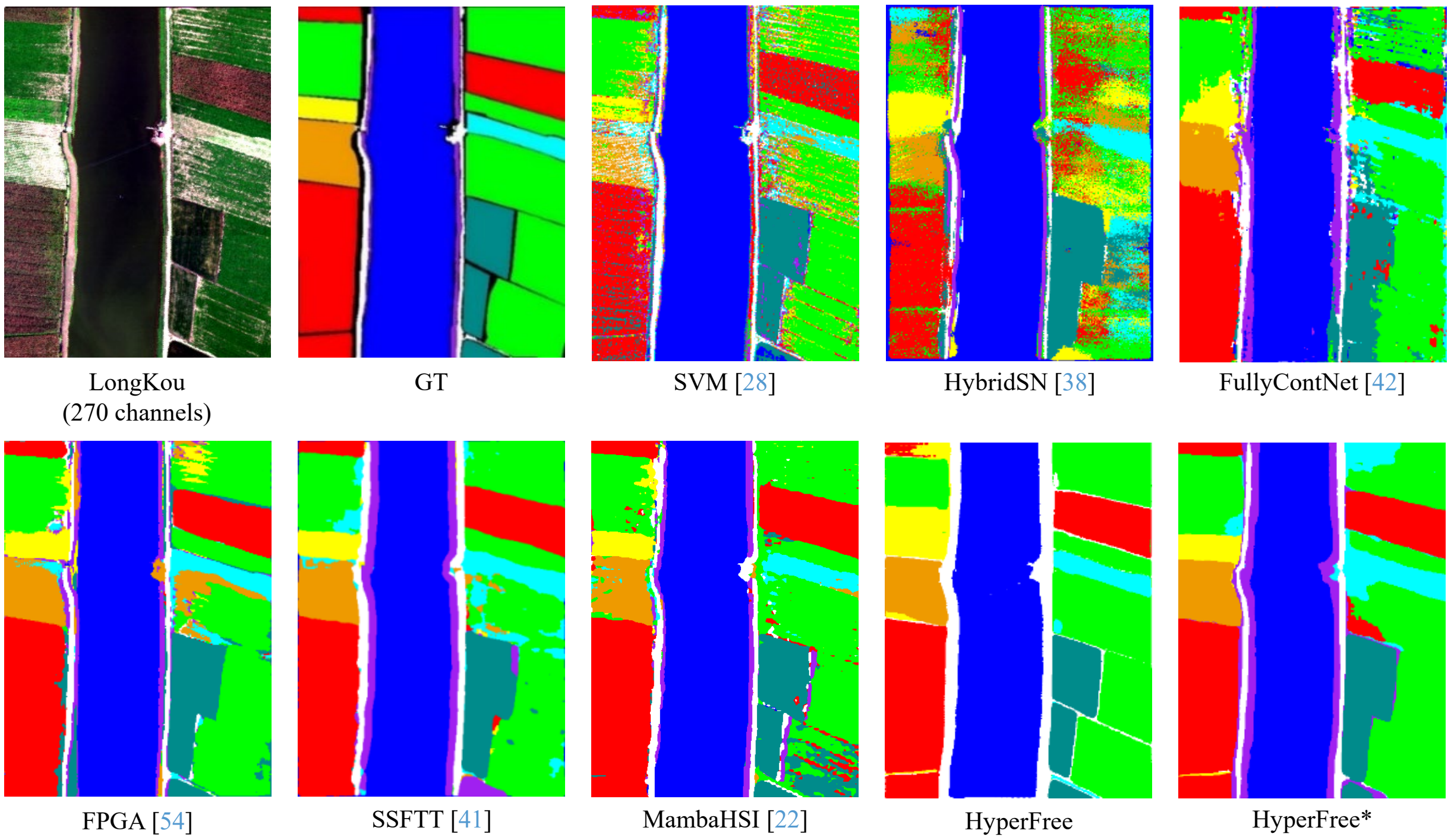

Figure 3. Qualitative comparison results on LongKou dataset of HC task, where HyperFree* represents the tuning version.

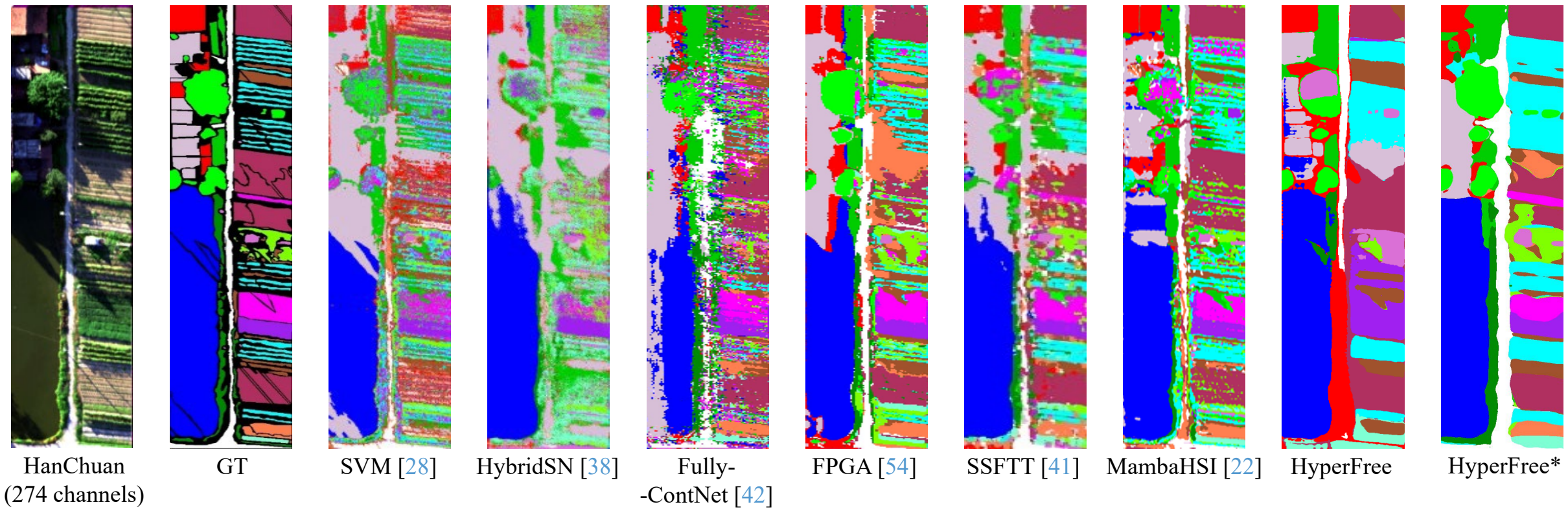

Figure 4. Qualitative comparison results on HanChuan dataset of HC task, where HyperFree* represents the tuning version.

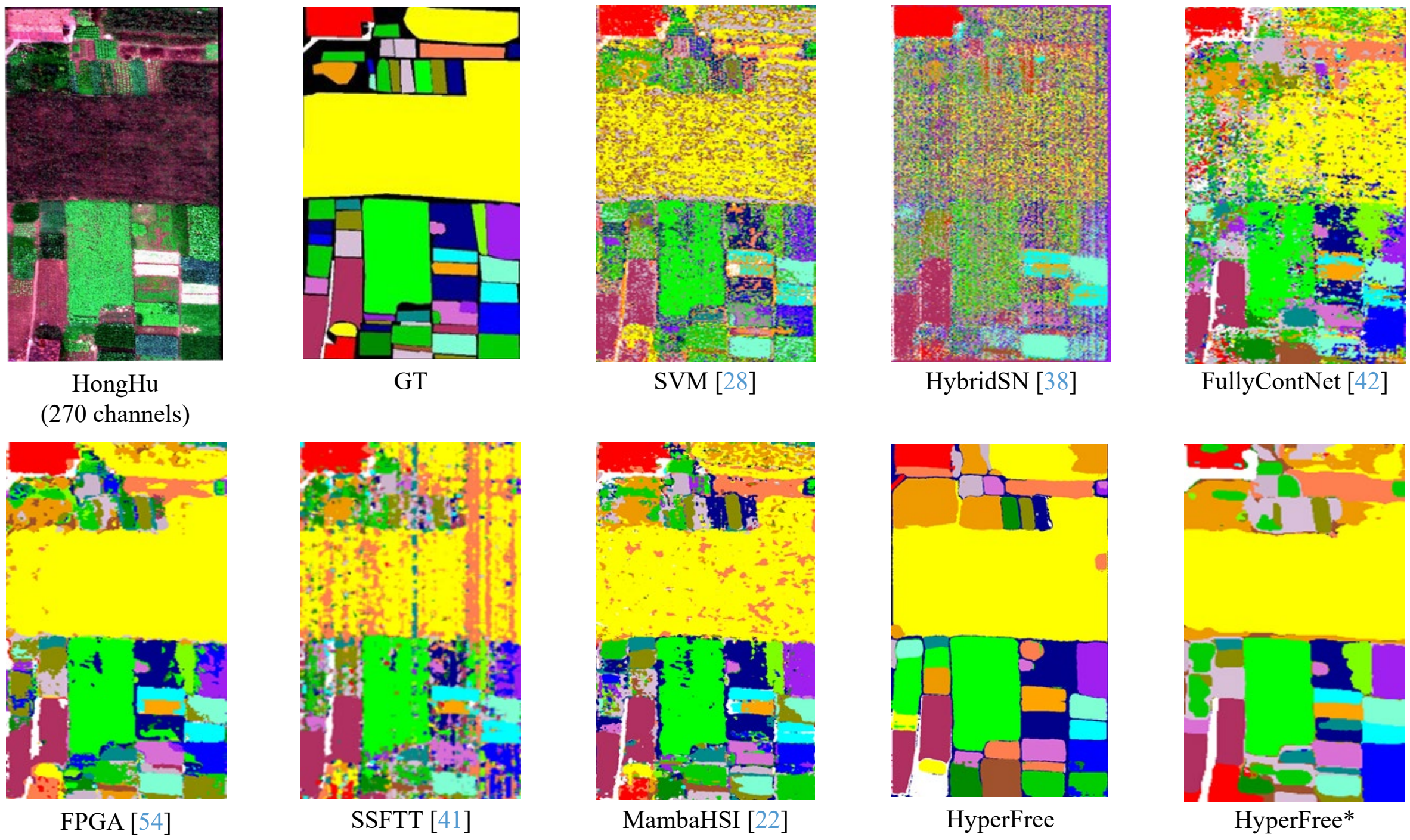

Figure 5. Qualitative comparison results on HongHu dataset of HC task, where HyperFree* represents the tuning version.

Table 5. Quantitative comparison results on HOCC task in tuning manner, where HyperFree* represents the tuning version and blue numbers indicate the metric ranking.

| Dataset | Metric | OCSVM [39] (5 shot) | nnPU [16] (5 shot) | BSVM [33] (5 shot) | PAN [13] (5 shot) | OC Loss [53] (5 shot) | T-HOneCls [52] (5 shot) | HyperFree* (5 shot) |
|---|---|---|---|---|---|---|---|---|
| HongHu [55] | $F_1$ | 26.33 | 19.13 | 34.82 | 63.69 | 54.73 | 72.52 | 91.24(1st) |
| | P | 56.43 | 19.72 | 50.79 | 75.00 | 58.26 | 46.52 | 92.77(1st) |
| | R | 24.02 | 18.58 | 45.29 | 64.27 | 54.34 | 92.35 | 89.90(2nd) |
| XiongAn [50] | $F_1$ | 18.31 | 1.76 | 26.30 | 46.34 | 43.08 | 41.34 | 66.89(1st) |
| | P | 39.83 | 2.85 | 23.82 | 47.13 | 47.50 | 32.87 | 61.94(1st) |
| | R | 16.08 | 1.98 | 57.83 | 53.32 | 47.61 | 60.38 | 75.74(1st) |

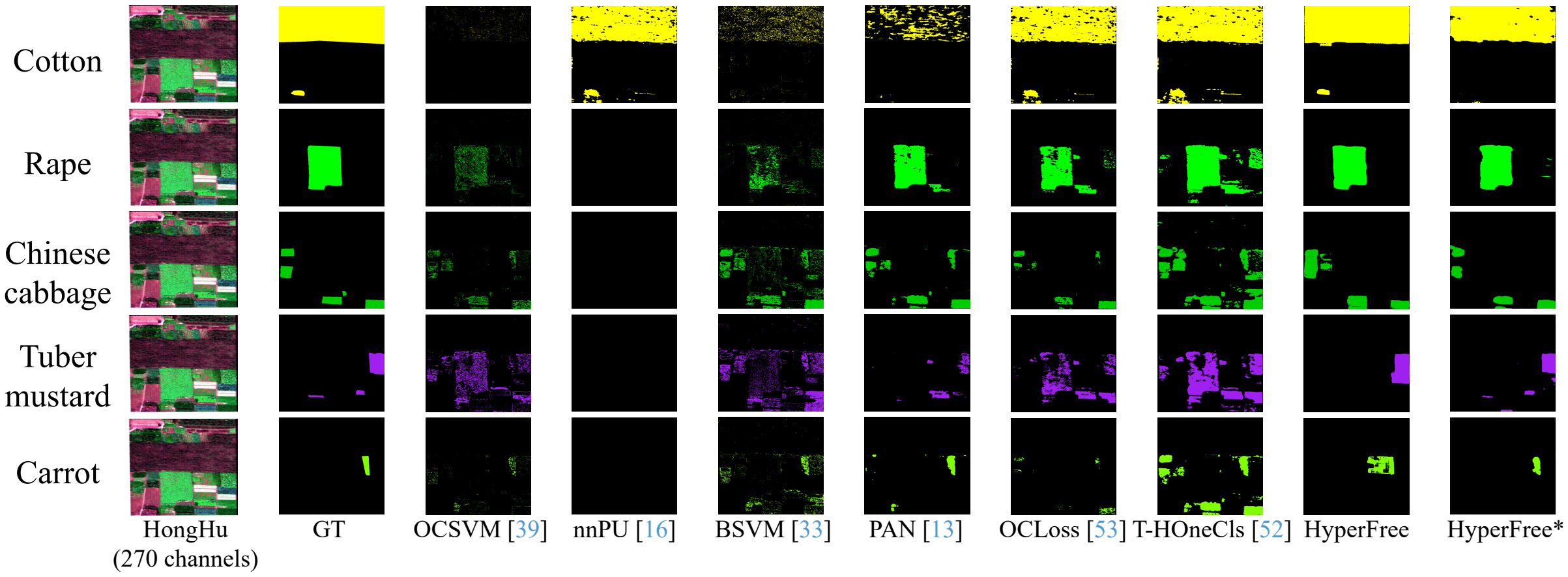

Figure 6. Qualitative comparison results on HongHu dataset of HOCC task, where HyperFree* represents the tuning version.

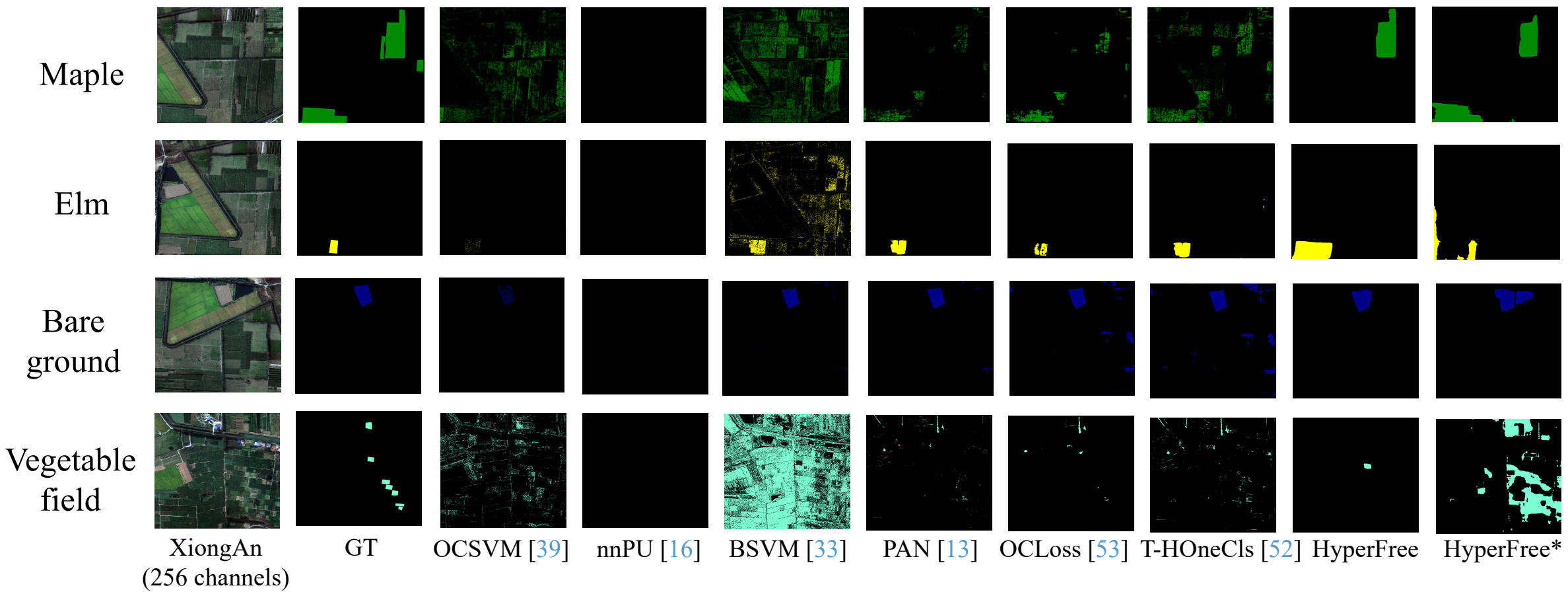

Figure 7. Qualitative comparison results on XiongAn dataset of HOCC task, where HyperFree* represents the tuning version.

Table 6. Quantitative comparison results on HTD task in tuning manner, where HyperFree* represents the tuning version and blue numbers indicate the metric ranking.

| Dataset | Metric | ACE [17] (1 shot) | CEM [4] (1 shot) | GLRT [26] (1 shot) | MF [27] (1 shot) | HTD-IRN [40] (1 shot) | TSTTD [14] (1 shot) | HyperFree* (1 shot) |
|---|---|---|---|---|---|---|---|---|
| Airport [2] | $AUC_{(D,F)}$ | 0.9794 | 0.9603 | 0.9801 | 0.9916 | 0.9745 | 0.9929 | 0.9937(1st) |
| | $AUC_{ODP}$ | 1.5853 | 1.2829 | 1.5798 | 1.6968 | 1.4484 | 1.6592 | 1.5945(3rd) |
| Cri [51] | $AUC_{(D,F)}$ | 0.9735 | 0.9893 | 0.9737 | 0.9891 | 0.9975 | 0.9987 | 0.9976(2nd) |
| | $AUC_{ODP}$ | 1.2015 | 1.4506 | 1.2 | 1.4575 | 1.3995 | 1.6103 | 1.4821(2nd) |

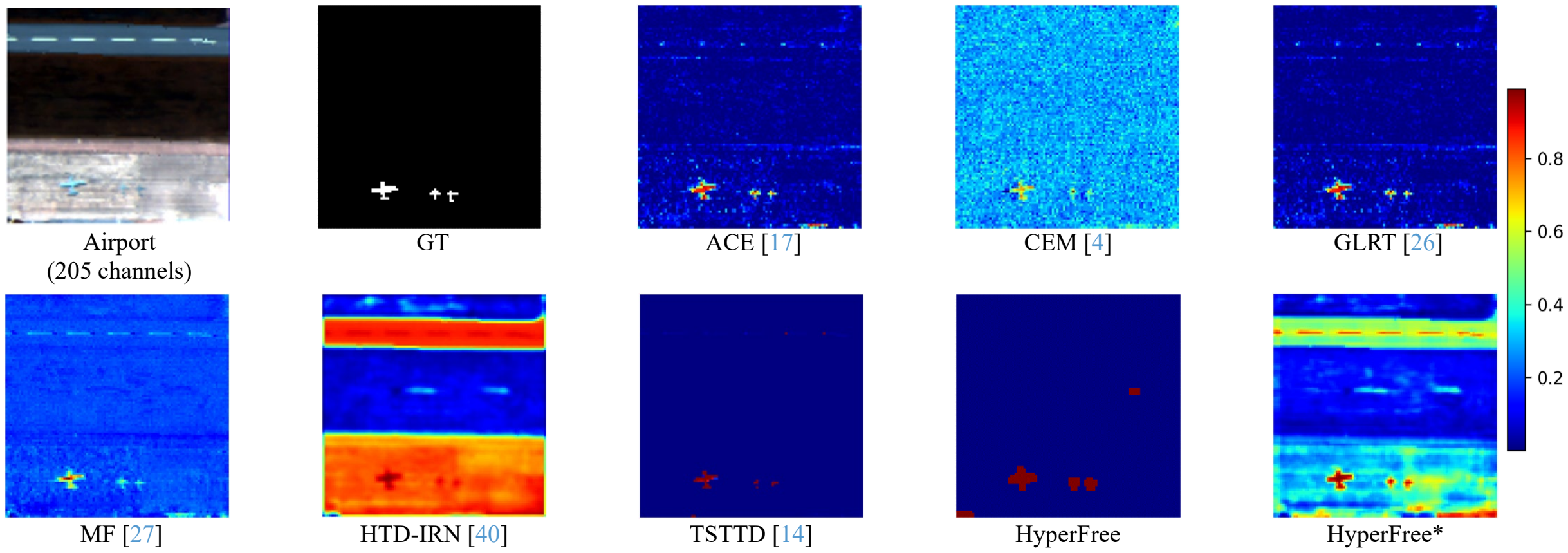

Figure 8. Qualitative comparison results on Airport-4 dataset of HTD task, where HyperFree* represents the tuning version.

Table 7. Quantitative comparison results on HAD task in tuning manner, where HyperFree* represents the tuning version and blue numbers indicate the metric ranking.

| Dataset | Metric | RXD [37] | CRD[20] | ADLR [36] | LRASR [49] | Auto-AD [43] | TDD [18] | HyperFree* |
|---|---|---|---|---|---|---|---|---|
| Beach-1 [2] | $AUC_{(D,F)}$ | 0.9815 | 0.9471 | 0.4515 | 0.7461 | 0.9574 | 0.9842 | 0.9973(1st) |
| | $AUC_{ODP}$ | 1.2557 | 0.9785 | 0.561 | 0.8526 | 1.1273 | 1.1383 | 1.7862(1st) |
| Beach-2 [2] | $AUC_{(D,F)}$ | 0.909 | 0.8544 | 0.7976 | 0.8225 | 0.9485 | 0.9627 | 0.9715(1st) |
| | $AUC_{ODP}$ | 1.0177 | 0.867 | 0.9064 | 0.828 | 1.0097 | 1.1688 | 1.3900(1st) |

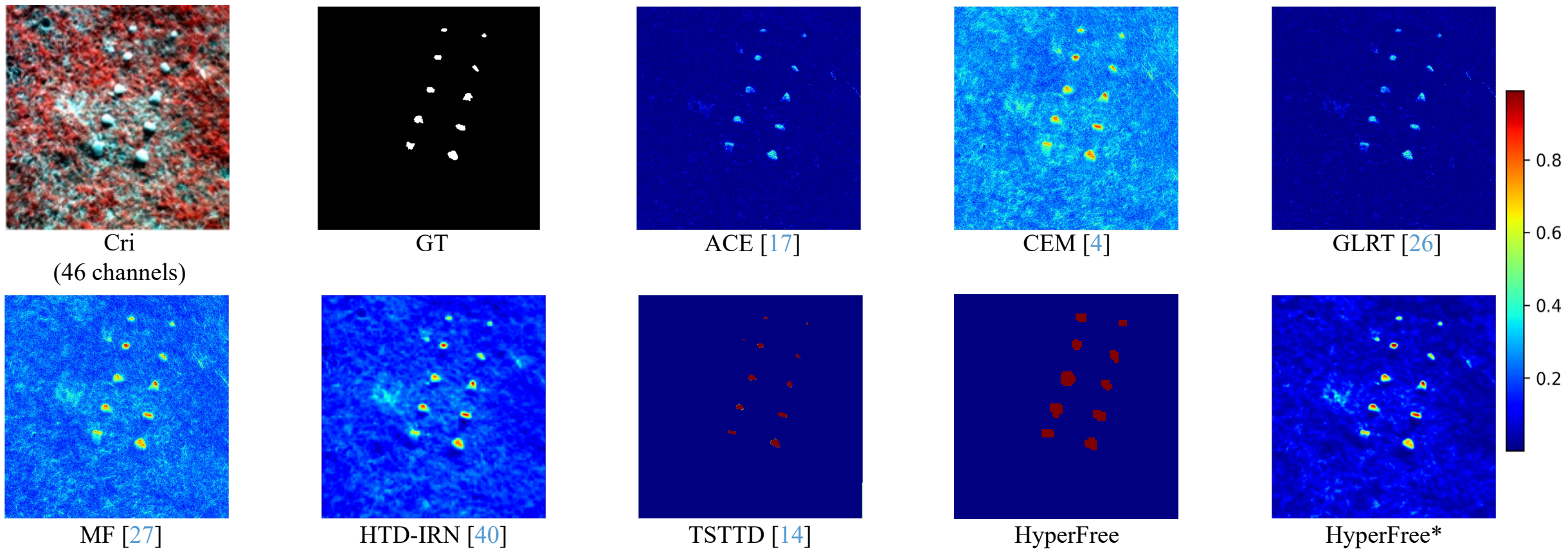

Figure 9. Qualitative comparison results on Cri dataset of HTD task, where HyperFree* represents the tuning version.

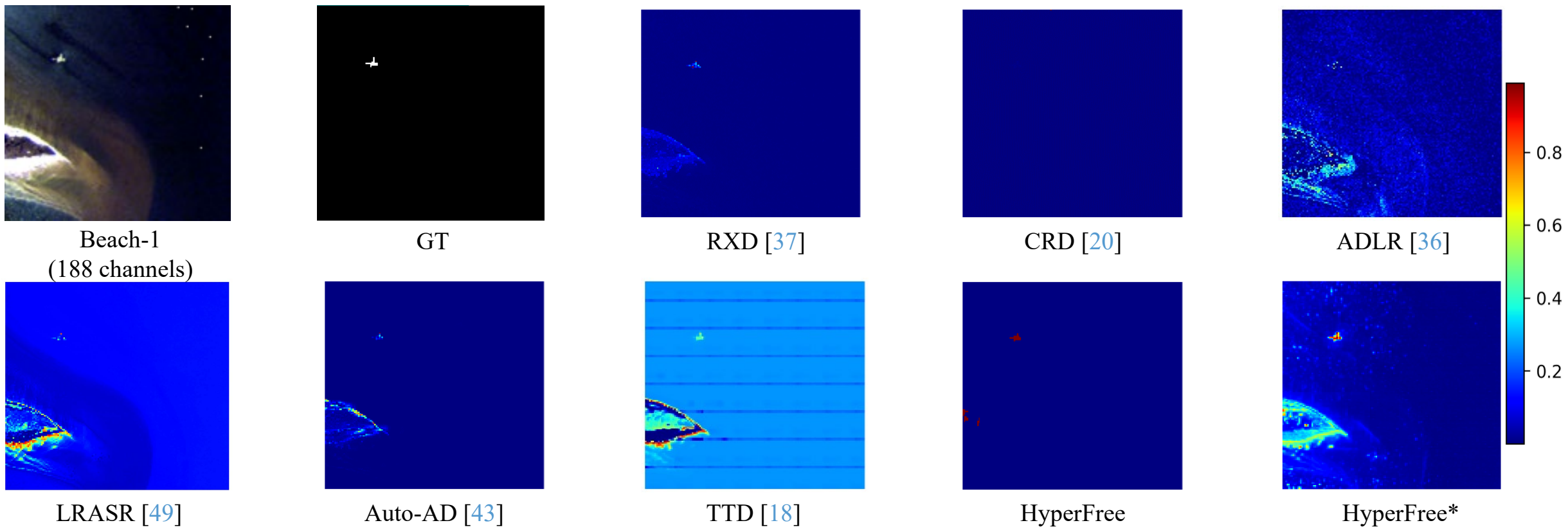

Figure 10. Qualitative comparison results on Beach-1 dataset of HAD task, where HyperFree* represents the tuning version.

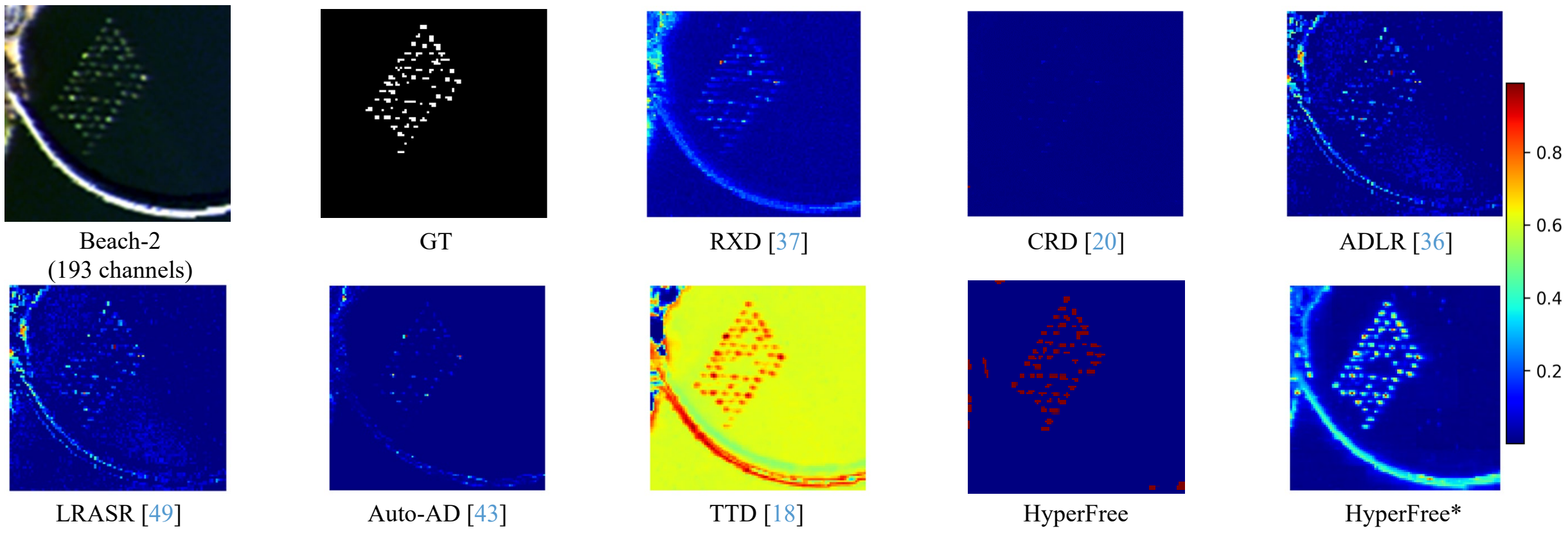

Figure 11. Qualitative comparison results on Beach-2 dataset of HAD task, where HyperFree* represents the tuning version.

Table 8. Quantitative comparison results on HCD task in tuning manner, where HyperFree* represents the tuning version and blue numbers indicate the metric ranking.

| Dataset | Metric | FC-EF [7] (5 shot) | FC-Sc [7] (5 shot) | FC-Sd [7] (5 shot) | ML-EDAN [34] (5 shot) | BIT [5] (5 shot) | SST-Former [45] (5 shot) | HyperFree* (5 shot) |
|---|---|---|---|---|---|---|---|---|
| Hermiston [12] | IoU | 37.29 | 37.76 | 48.73 | 32.52 | 52.57 | 53.61 | 61.58(1st) |
| | $F_1$ | 54.32 | 54.82 | 65.52 | 49.08 | 68.91 | 69.8 | 76.22(1st) |
| River [12] | IoU | 41.68 | 45.22 | 45.34 | 39.15 | 21.26 | 40.96 | 47.16(1st) |
| | $F_1$ | 58.84 | 62.28 | 62.39 | 56.28 | 35.07 | 58.12 | 64.09(1st) |

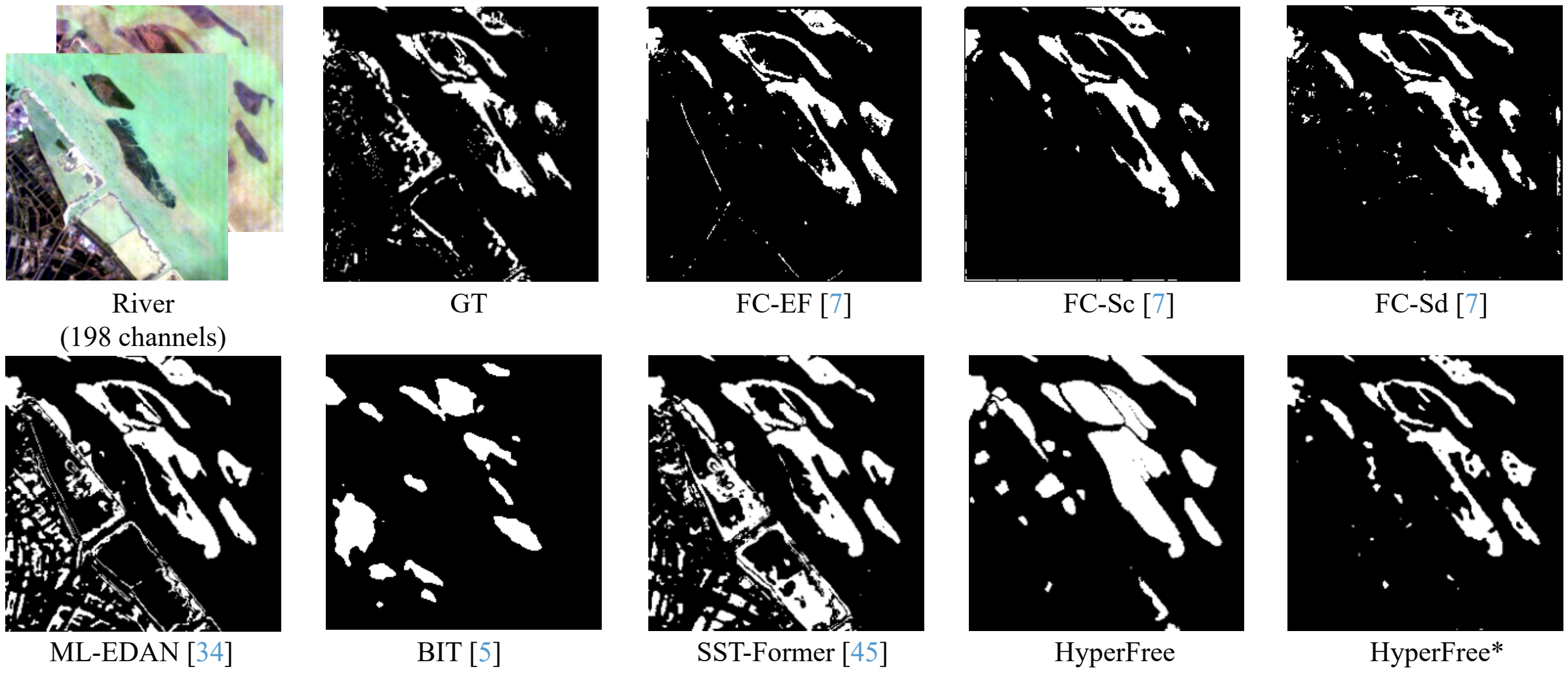

Figure 12. Qualitative comparison results on River dataset of HCD task, where HyperFree* represents the tuning version.

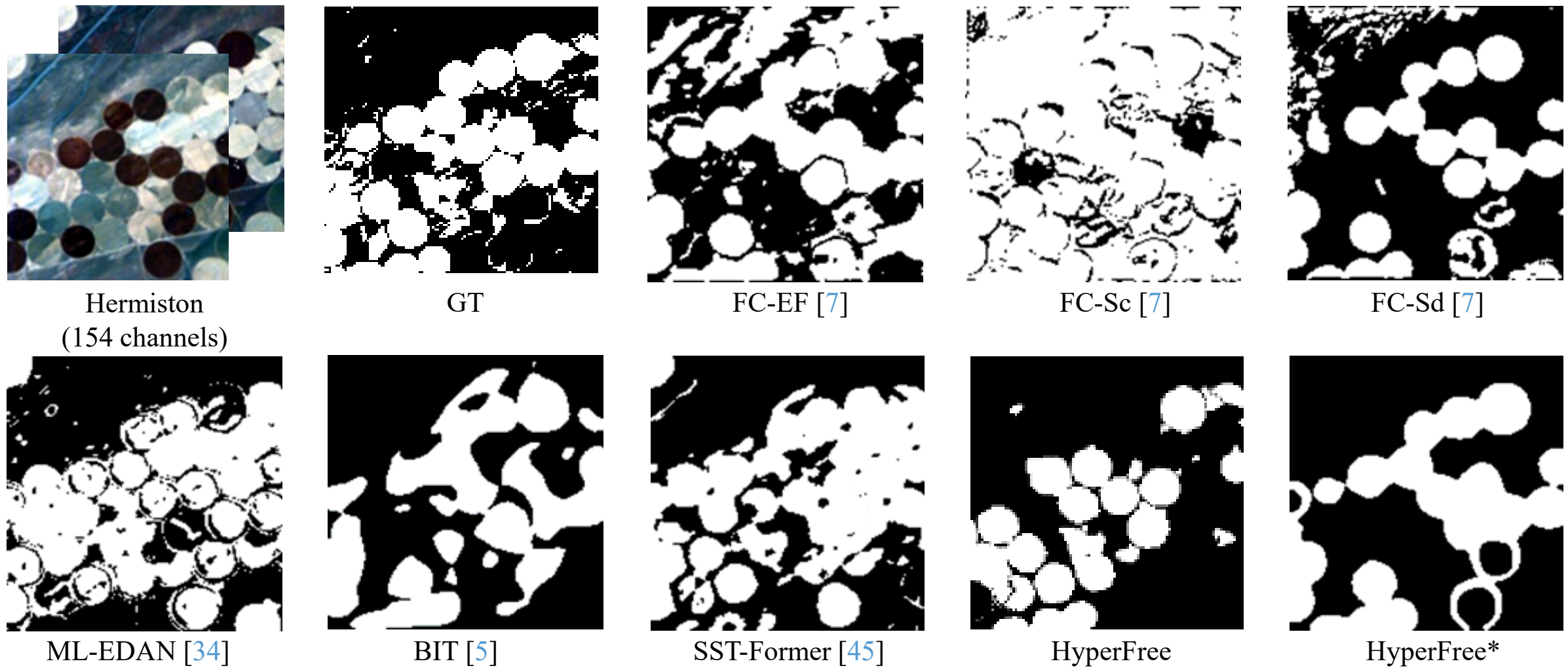

Figure 13. Qualitative comparison results on Hermiston dataset of HCD task, where HyperFree* represents the tuning version.

Table 9. Quantitative comparison results on HD task in tuning manner, where HyperFree* represents the tuning version and blue numbers indicate the metric ranking.

| Dataset | Metrics | NGMee [9] | $LRTFL_0$ [47] | E-3DTV [32] | QRNN3D [46] | DS2DP[29] | SST [19] | HyperFree* |
|---|---|---|---|---|---|---|---|---|
| Washington D.C. [3] | PSNR | 23.89 | 25.58 | 25.97 | 27.79 | 27.31 | 28.1 | 28.49(1st) |
| | SSIM | 0.872 | 0.907 | 0.921 | 0.945 | 0.937 | 0.989 | 0.990(1st) |
| | SAM(°) | 14.89 | 11.35 | 8.772 | 7.563 | 7.735 | 6.343 | 6.251(1st) |

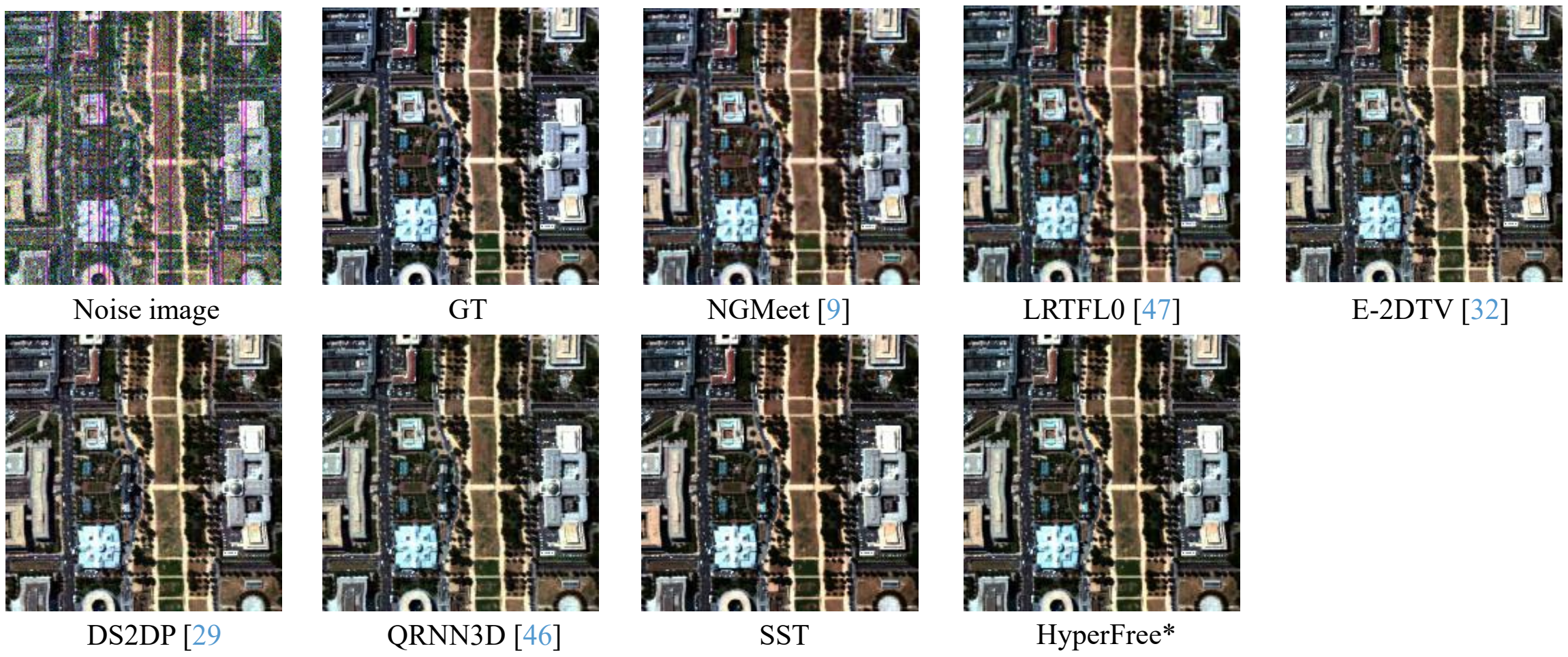

Figure 14. Qualitative comparison results on Washington D.C. dataset of HD task, where HyperFree* represents the tuning version.

Table 10. Quantitative comparison results on HU task in tuning manner, where HyperFree* represents the tuning version and blue numbers indicate the metric ranking.

| Dataset | Metric | VCA-FCLS [10, 30] | SGSNMF [44] | uDAS [35] | CNNAEU [31] | CyCU-Net [8] | GSUU [6] | HyperFree* |
|---|---|---|---|---|---|---|---|---|
| Urban [15] | SAD | 0.3859 | 0.4442 | 0.6498 | 0.5364 | 0.2750 | 0.1645 | 0.0446(1st) |
| | RMSE | 0.1061 | 0.0973 | 0.1009 | 0.0392 | 0.1597 | 0.1188 | 0.0170(1st) |

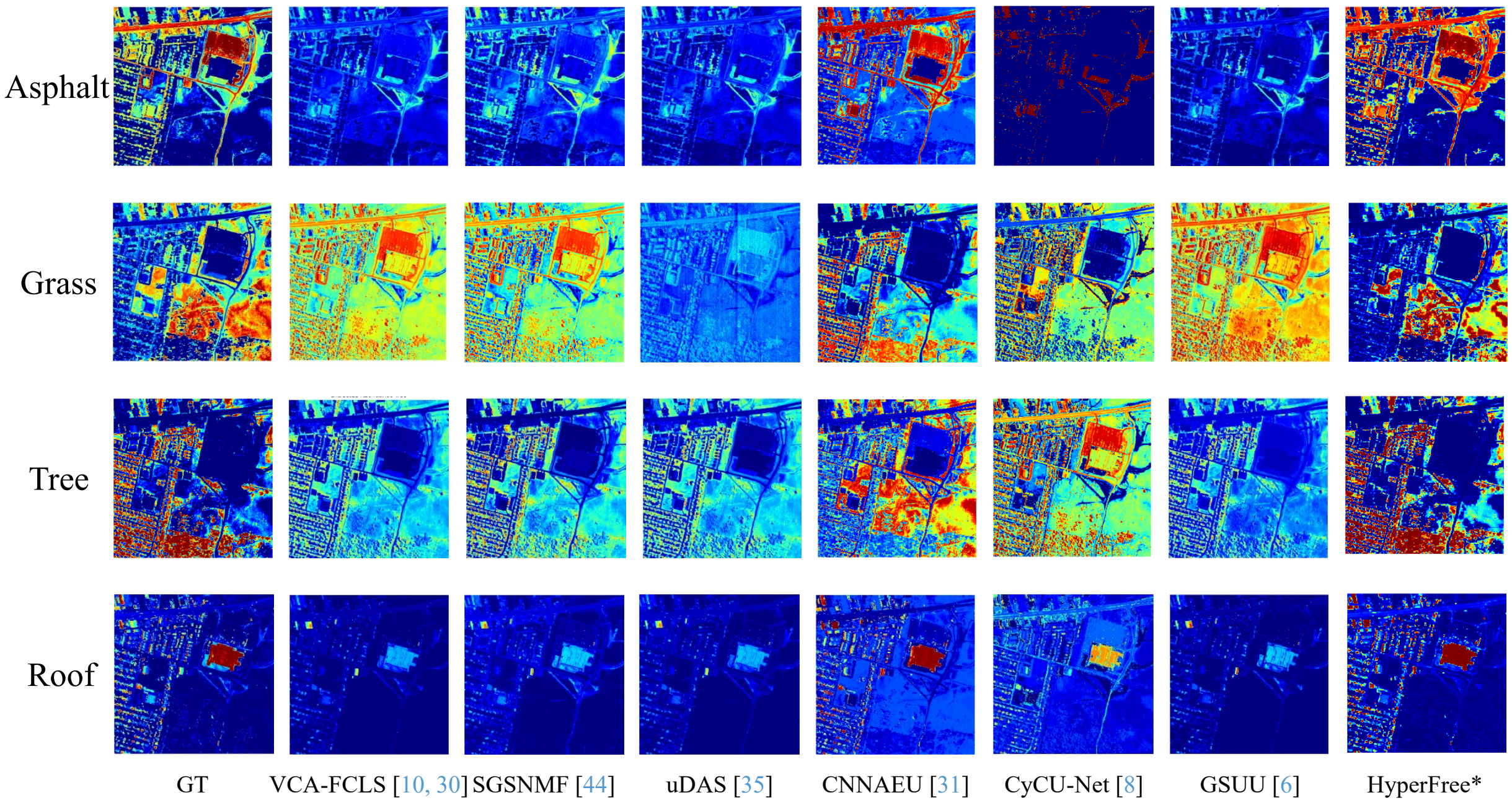

Figure 15. Qualitative comparison results on Urban dataset of HU task, where HyperFree* represents the tuning version.

Table 11. Quantitative comparison results on HOT task in tuning manner, where HyperFree* represents the tuning version and blue numbers indicate the metric ranking.

| Data | Metrics | BAENet [23] | MHT [48] | SiamHYPER [25] | SEE-Net [24] | SiamBAG [21] | TSCFW [11] | HyperFree* |
|---|---|---|---|---|---|---|---|---|
| HOTC 2023 [1] | AUC | 0.496 | 0.465 | 0.564 | 0.499 | 0.508 | 0.476 | 0.576(1st) |
| | DP | 0.757 | 0.733 | 0.778 | 0.737 | 0.736 | 0.708 | 0.796(1st) |
| | FPS | 0.8 | 0.5 | 29.8 | 16.8 | 14.1 | 4.1 | 16.4(3rd) |

**Name**: VIS_Coke **Attributes**: BC, IPR, OPR, FM, SV

**Name**: NIR_Car59 **Attributes**: OCC, SV, FM

**Name**: RedNIR_Dice2 **Attributes**: LR, BC, OCC

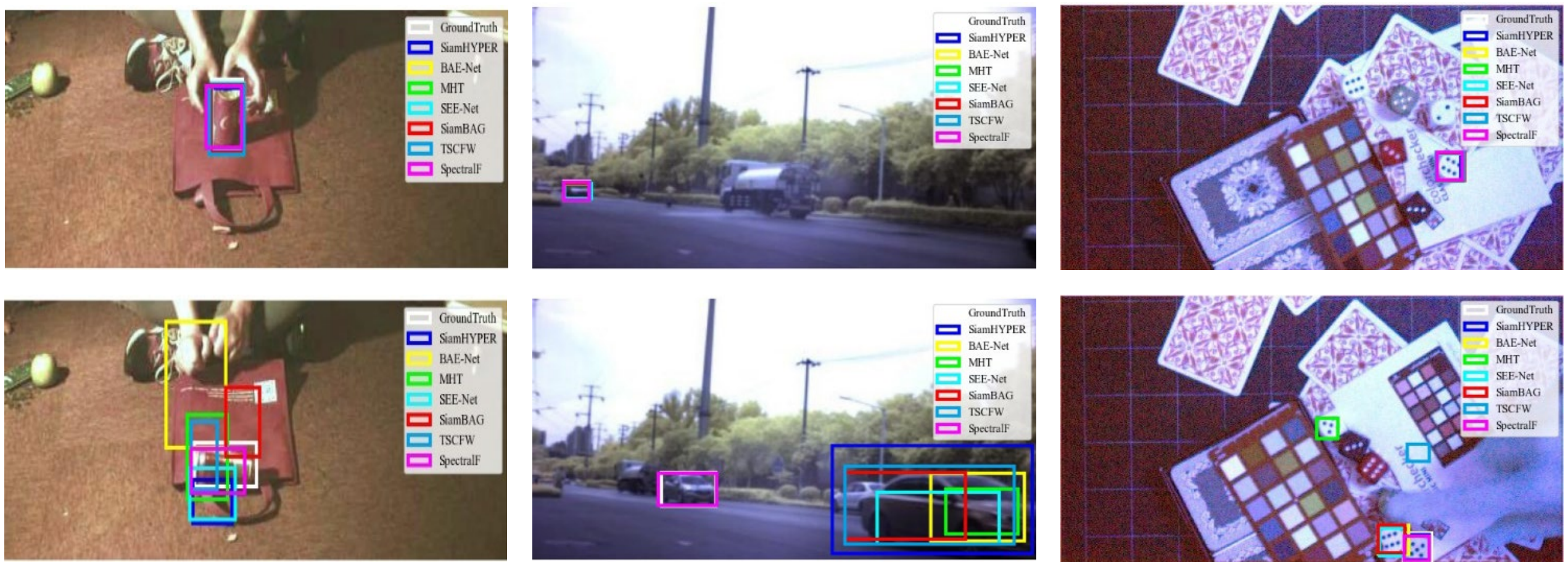

Figure 16. Qualitative comparison results on HOCT 2023 dataset of HOT task, where HyperFree* represents the tuning version.

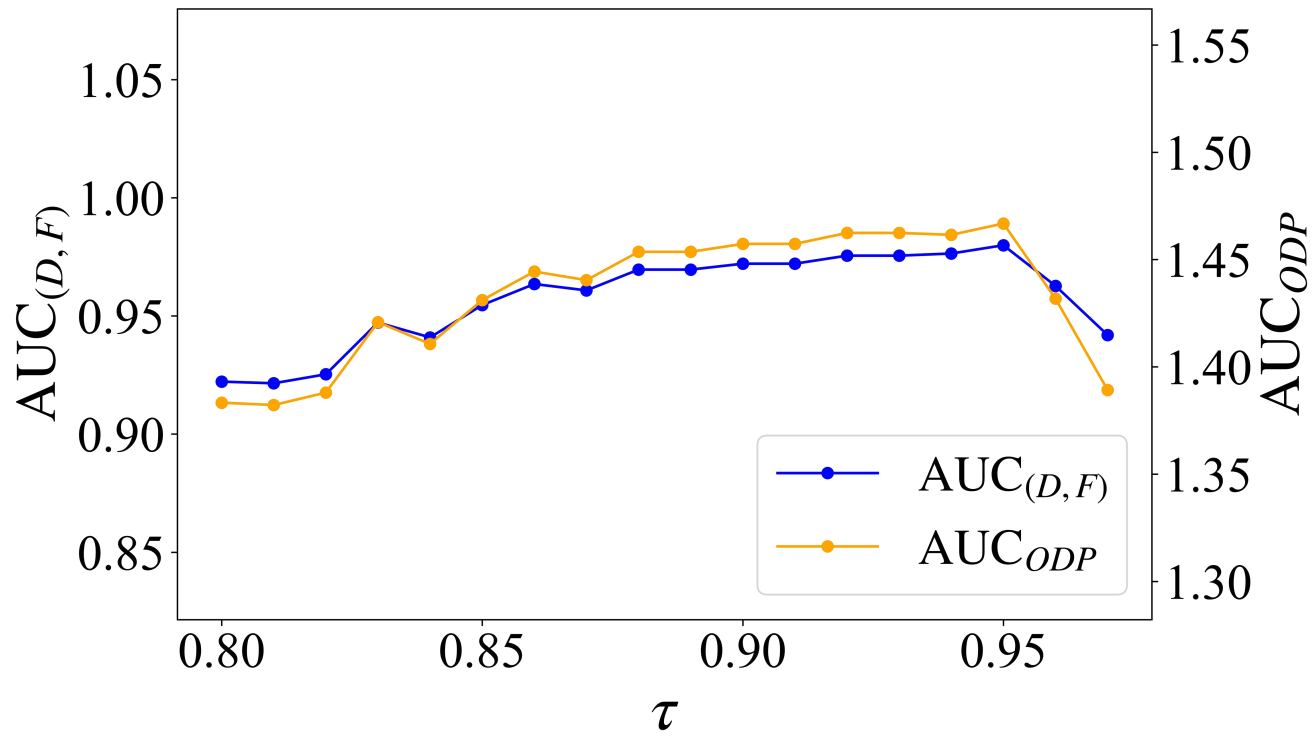

(a) HTD on Cri Dataset

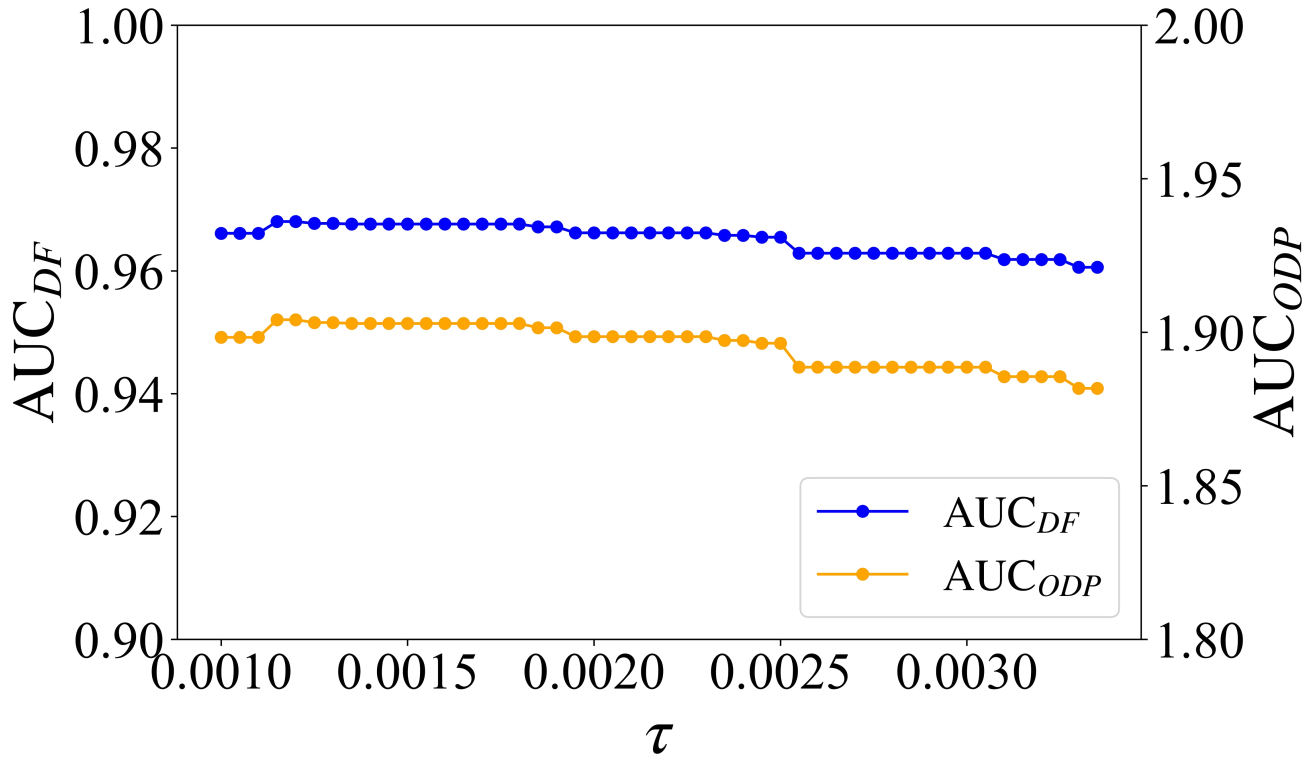

(b) HAD on Beanch Dataset

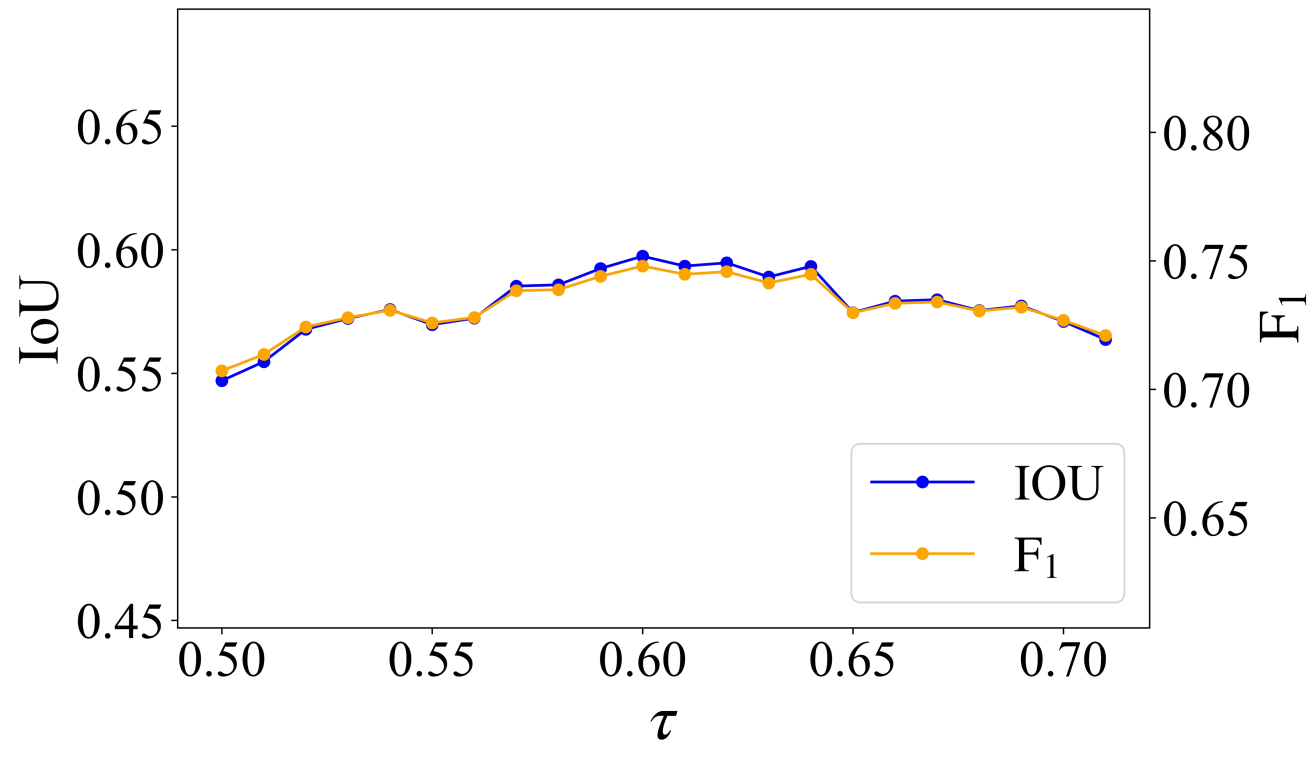

(c) HCD on Hermiston Dataset

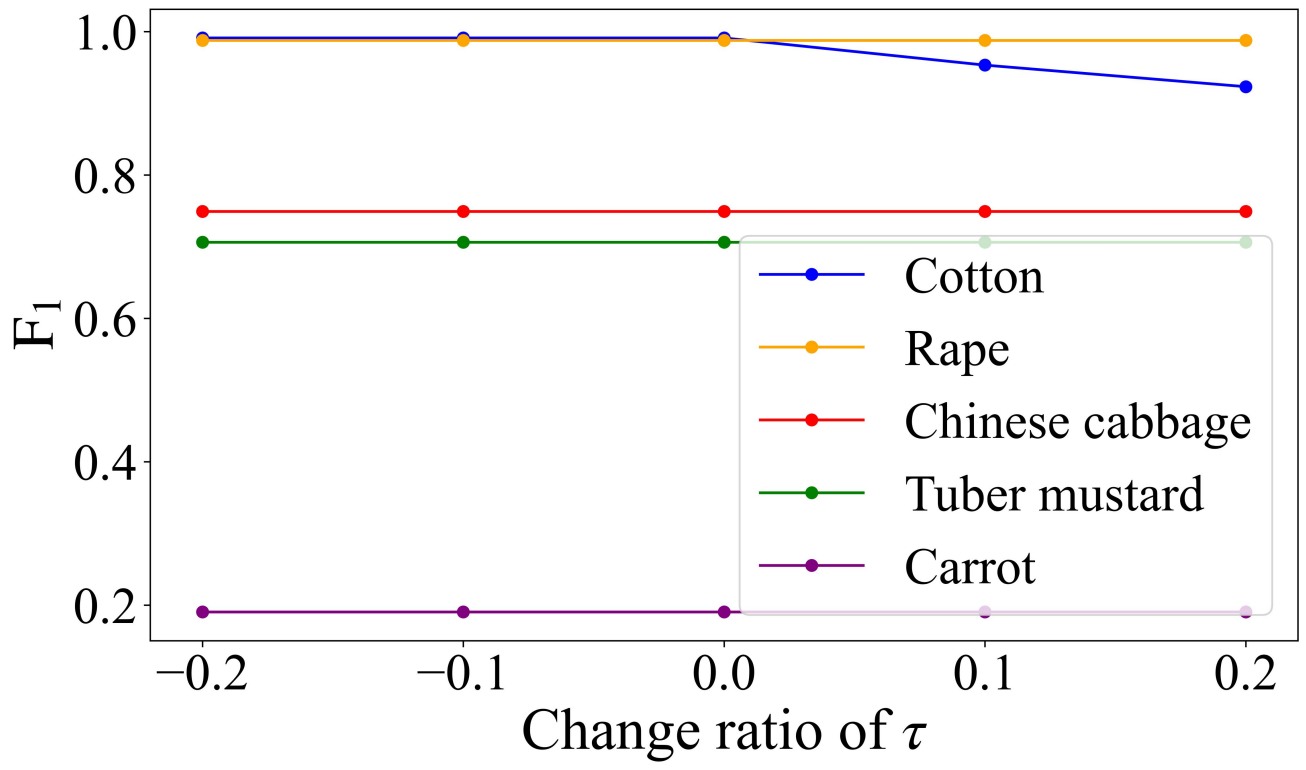

(d) HOCC on HongHu Dataset

Figure 17. Sensitivity analysis of the hyperparameter $\tau$ on the model performance.